\documentclass[acmsmall]{acmart}

\setcopyright{acmcopyright}
\copyrightyear{2024}
\acmYear{2024}
\acmDOI{XXXXXXX.XXXXXXX}

\usepackage[linesnumbered,ruled,vlined]{algorithm2e}
\usepackage{multirow}
\usepackage{booktabs}
\usepackage{caption}
\SetAlCapFnt{\small}
\SetAlCapNameFnt{\small}

\usepackage{hyperref}

\begin{document}

\title{Extracting Human Attention through Crowdsourced Patch Labeling}

\author{Minsuk Chang}
\email{jangsus1@snu.ac.kr}
\affiliation{%
  \institution{Seoul National University}
  \city{Seoul}
  \country{Republic of Korea}
}

\author{Seokhyeon Park}
\email{shpark@hcil.snu.ac.kr}
\affiliation{%
  \institution{Seoul National University}
  \city{Seoul}
  \country{Republic of Korea}
}

\author{Hyeon Jeon}
\email{hj@hcil.snu.ac.kr}
\affiliation{%
  \institution{Seoul National University}
  \city{Seoul}
  \country{Republic of Korea}
}

\author{Aeri Cho}
\email{archo@hcil.snu.ac.kr}
\affiliation{%
  \institution{Seoul National University}
  \city{Seoul}
  \country{Republic of Korea}
}

\author{Soohyun Lee}
\email{shlee@hcil.snu.ac.kr}
\affiliation{%
  \institution{Seoul National University}
  \city{Seoul}
  \country{Republic of Korea}
}

\author{Jinwook Seo}
\email{jseo@snu.ac.kr}
\affiliation{%
  \institution{Seoul National University}
  \city{Seoul}
  \country{Republic of Korea}
}

\renewcommand{\shortauthors}{Chang et al.}


\begin{abstract}
   In image classification, a significant problem arises from bias in the training datasets. When a dataset contains only specific types of images, the classifier begins to rely on shortcuts - simplistic and erroneous rules for decision-making. This leads to high performance on the training dataset but inferior results on new, varied images, as the classifier's generalization capability is reduced. For example, if the images labeled as \textit{mustache} consist solely of male figures, the model may inadvertently learn to classify images by gender rather than the presence of a mustache. A common approach to mitigate such biases is to direct the model's attention toward the target object's location, which is marked using bounding boxes or polygons for accurate annotation. However, collecting such annotations requires substantial time and human effort. In this research, we propose a novel patch-labeling method that integrates AI assistance with crowdsourcing to capture human attention from images, which can be a viable solution for mitigating bias. Our method consists of two steps. First, we extract the approximate location of a target using a pre-trained saliency detection model, supplemented by human verification for accuracy. Then, we determine the human-attentive area in the image by iteratively dividing the image into smaller patches and employing crowdsourcing to ascertain whether each patch can be classified as the target object. We demonstrated the effectiveness of our method in mitigating bias, as evidenced by improved classification accuracy and the refined focus of the model. Also, crowdsourced experiments validate that our method collects human annotation up to 3.4 times faster than annotating object locations with polygons, significantly reducing the need for human resources. We conclude the paper by discussing the advantages of our method in a crowdsourcing context, particularly focusing on aspects of human errors and accessibility.
 
\end{abstract}



\begin{teaserfigure}
  \includegraphics[width=\textwidth]{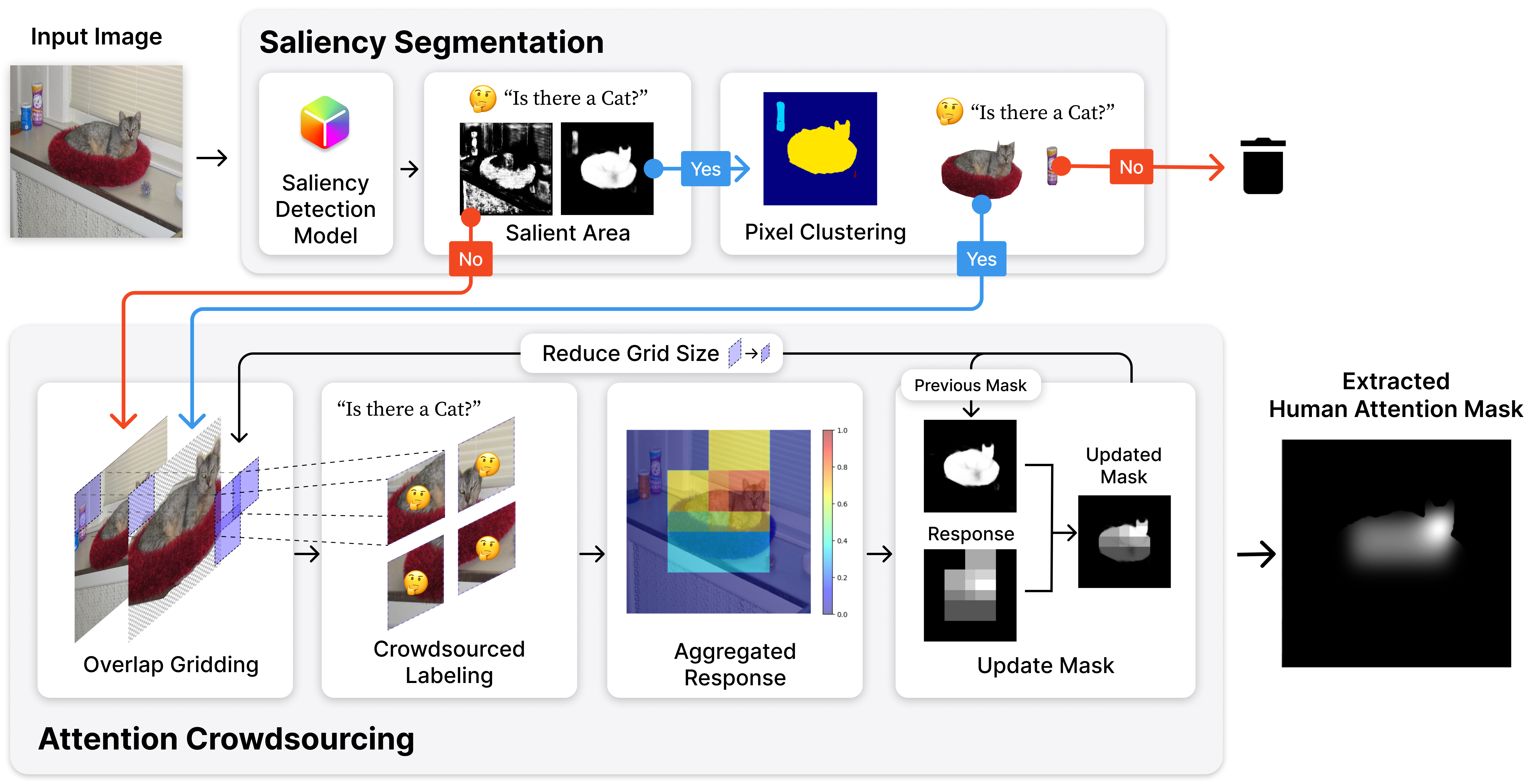}
  \caption{We start by processing an image through \textbf{Saliency Segmentation} to identify potential object locations. Next, we precisely locate the object in the \textbf{Attention Crowdsourcing} phase. This two-step approach generates a map highlighting where humans will most likely focus their attention during classification.}
  \Description{Human Attention Extraction process from the input image to final attention mask.}
  \label{fig:teaser}
\end{teaserfigure}

\received{15 January 2024}

\maketitle

\section{Introduction} \label{Introduction}
People train image classifiers with labeled datasets to predict the class of unseen images. However, datasets limited to specific types of images can lead to a decrease in test accuracy for these trained classifiers due to a diminished capacity for generalizability. This problem, so-called bias in image datasets, arises from the frequent co-occurrence of objects within these images, forcing the model to adopt shortcuts - rules that are easy to learn but ultimately incorrect \cite{COCOBias, multipleShortcuts}. For example, in the COCO \cite{coco} dataset, the \textit{Horse} class predominantly features images with grassfield backgrounds, lacking variety in terms of location diversity. The CelebA \cite{celeba} dataset has gender biases; a notable example is in attributes such as \textit{Wearing Lipstick} \cite{CelebaBias}, where the dataset exclusively includes images of female celebrities wearing lipsticks. When an image classifier is trained on these biased datasets, the model's predictions may also become biased, disproportionately paying more attention to the bias factors. As a result, the models may suffer from a significant reduction in accuracy when they are deployed in the wild, where more diverse images exist.
\par
There were various approaches to prevent the model from learning bias by providing additional clues. For example, one approach fixed the biased image classifier with a human-in-the-loop system \cite{guide}, where humans directly guide the model's attention. Another approach used image segmentation datasets that include the boundaries of the target object, compelling the model to concentrate on the appropriate object \cite{segmentation}. Other work utilized eye-tracking devices to capture the human gaze patterns for model training \cite{SaliencyClassification}. However, these methods often require substantial human effort in manually collecting boundary annotations or recording gaze with eye trackers. Also, directly guiding the biased models takes several hours, which becomes a severe burden.
\par
To address this challenge, we introduce a patch-labeling method designed to efficiently extract human attention from an image with significantly less human effort. The entire pipeline is depicted in~\autoref{fig:teaser}. It starts with the \textbf{Saliency Segmentation} step, which employs a salient object detection model to identify key regions within an image. Acknowledging that these regions might not always encompass the target object, human verification is incorporated to confirm the relevance of these identified areas. We grouped adjacent pixels in the saliency map to form object clusters, which are then rapidly validated by users, effectively filtering out non-essential regions. Subsequently, we apply the \textbf{Attention Crowdsourcing} step where the image is divided into small, overlapping grid patches. Crowd workers evaluate each patch to determine the presence of the target object. In this step, we only show patches rather than full images to ensure that information is gathered solely from specific regions, thereby removing unintended bias from other regions. The final step involves aggregating the responses from the workers to generate a comprehensive human attention mask.
\par
Our empirical evaluation showed that our method can effectively generate a practical annotation that represents human attention, thereby help mitigating model bias. Our crowdsourced experiments demonstrated that our method is more time-efficient in obtaining the annotation than marking the object boundary with a polygon-drawing tool \cite{segmentation}. Furthermore, we observed that our method has low variance in the labeling time compared to the polygon-drawing approach, which indicates that our method maintains constant performance, even in a crowdsourcing context. Additional experiments with image classifiers revealed that the accuracy significantly increases when the model is trained to align its attention with our human attention mask. We also demonstrated that our model predicts with less confidence on images that solely contain the bias factors, compared to the baseline model trained without such guidance.
\par
Our adoption of the patch-labeling approach brings three key advantages to crowdsourced labeling. First, erroneous or malicious responses can be automatically filtered out by calculating the validity score of each submission. Secondly, it can offer enhanced parallelization capabilities; dividing the tasks into smaller image patches facilitates a more granular distribution of crowd workers. Theoretically, our method is expected to expedite the labeling task by almost 30 times compared to the polygon drawing approach. Lastly, our method can be easily deployed across various platforms and devices while demonstrating robustness against individual variances in labeling.

The contribution of our work is three-fold.
\begin{itemize}
    \item We present a novel method to extract the human attention mask by combining a salient object detection model and crowdsourced patch-labeling, which can be used to train an unbiased image classifier.

    \item We empirically proved that the human attention mask generated by our method can guide the model's attention to focus on the target object, making it robust toward bias factor with higher performance. Also, our crowdsourced experiments show that our approach is time- and cost-efficient to obtain compared to the polygon annotations.

    \item Our proposed method has three potential advantages in crowdsourced labeling: 1) resilience to potential human errors, 2) fine-grained parallelism to achieve faster task completion, and 3) a patch-labeling approach ensuring easy deployment and usage across various platforms and users.
    
\end{itemize}

\section{Related Work} \label{Related Work}
\begin{figure}[t]
  \centering
  \includegraphics[width=0.8\linewidth]{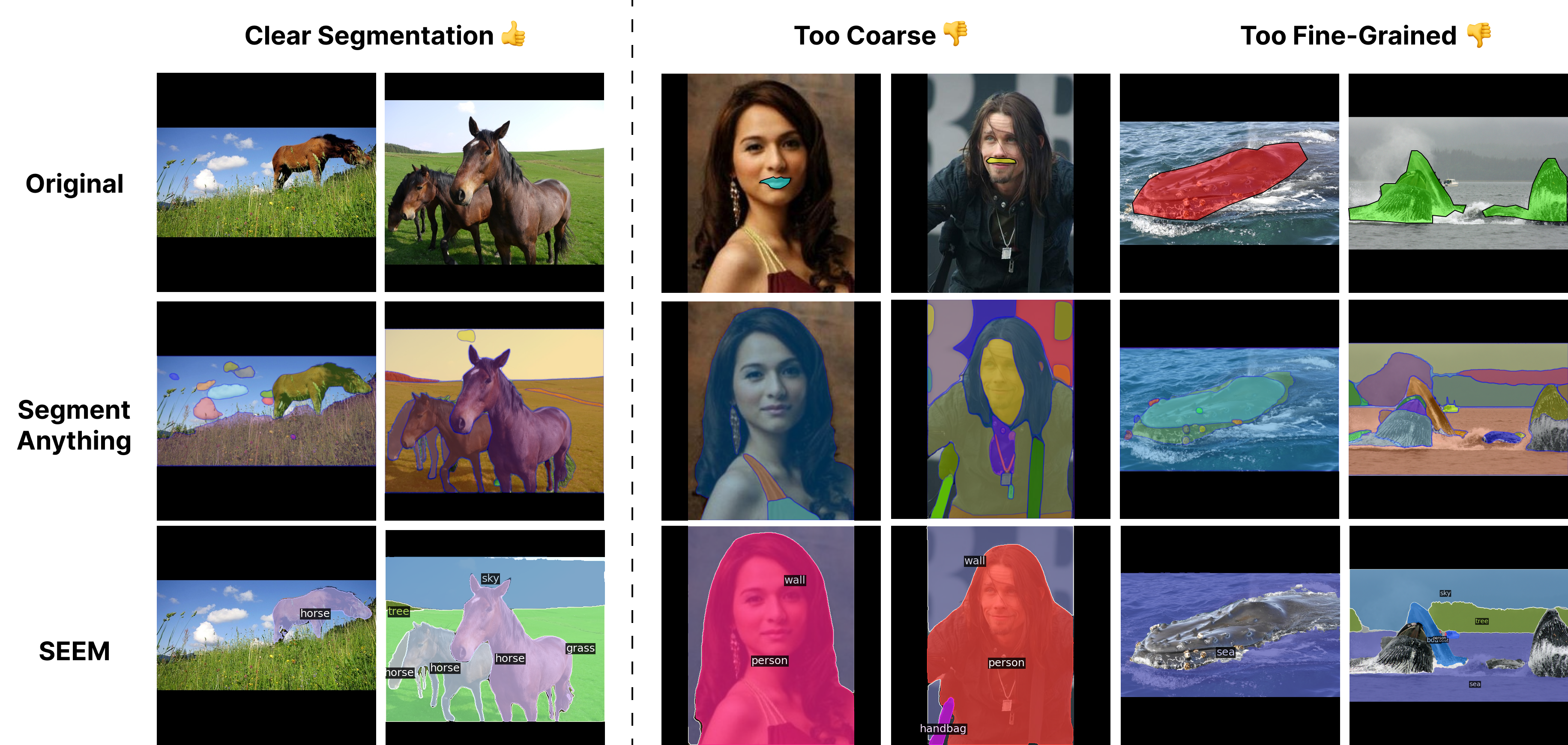}
  \caption{Identified problems of Segment Anything~\cite{SegmentationAnything} and SEEM~\cite{SegmentationSEEM} with images from CelebA and AwA2 dataset. Both models successfully segmented the animals in the typical background. However, they tend to merge the facial elements in the CelebA dataset, generating too coarse segments. Also, they produced scattered patches for marine animals in the AwA2 dataset or even failed to detect the animal.}
  \Description{}
  \label{fig:segmentAnything}

  \vspace{-3mm}
\end{figure}

This section examines key advancements in mitigating dataset bias, capturing human attention, and integrating AI in data labeling. These works highlight their collective impact on developing fair and efficient machine learning models, which our work pursues.

\subsection{Dataset Bias}
An analysis of dataset bias revealed that most datasets unintentionally include biases during collection~\cite{COCOBias}. Subsequent experiments identified various biases specific to certain datasets~\cite{CelebaBias, COCOBias, AwA2Bias}. For instance, an analysis of the CelebA dataset~\cite{CelebaBias} exposed a gender bias in the "Wearing Lipstick" category, leading models to base predictions on gender rather than the intended feature.
\par
There have been several proposals to mitigate such inherent biases in datasets. One approach involves using multiple datasets to enhance generalizability. This method separates the biases from the true features by assigning a learnable weight specifically for bias representation, while other weights represent actual features~\cite{BiasUnrolling}. Another technique involves reducing bias by identifying features common across different datasets, thus focusing on domain-invariant characteristics~\cite{BiasUnsupervised}. Additionally, some researchers have developed a deep neural network that focuses solely on features shared among datasets, which helps reduce dataset bias~\cite{BiasAdaptation}. However, these strategies have limitations, particularly because they depend on the availability of multiple datasets in similar domains, which is usually feasible only for common objects like transportation means or animals.
\par
Other strategies have been explored to prevent models from learning biases when trained on a single dataset, though these, too, come with certain drawbacks. One approach employs a regularization loss to prevent the model from acquiring biases but requires prior knowledge about them~\cite{BiasNormalization}. Another method eliminates bias based on feature representation within the classifier's latent space~\cite{BiasFeaturerepresentation}. However, this approach can be unreliable due to the opaque nature of deep learning models~\cite{DeeplearningUnreliable}. Additionally, there has been an effort to train an image reconstruction model that automatically removes bias factors from images. However, this technique has been observed to significantly degrade the quality and resolution of the reconstructed images~\cite{BiasDebiasModel}.
\par
To alleviate such limitations, our research uses more straightforward and reliable clues to guide the model, which is the extracted human attention. Unlike prior works, our method can extract human attention from any dataset following our algorithm in an explainable manner.

\subsection{Human Attention} \label{RelatedWork:HumanAttention}
The concept of human attention is multifaceted, leading to various methods for its capture and analysis. One technique involves using eye-tracking devices to collect gaze data, which has been applied to guide the attention of deep neural network (DNN) models~\cite{GazeSurvey}. Further developments in this area include capturing the focus areas within images for object classification purposes~\cite{SaliencyClassification} and gathering detailed gaze information for hierarchical classification tasks in specific domains like bird image taxonomies~\cite{EyeTrackingHierarchy}. Another study explored the objectivity of gaze-based saliency areas, proposing methods for gathering more unbiased data or synthesizing it~\cite{SaliencyImitator}. The relationship between human attention and DNN model attention has also been thoroughly investigated, especially under varying conditions such as spontaneous gaze or manual area selection~\cite{SaliencyVSModel}. However, these techniques are limited by the necessity for precise, often sensitive eye-tracking technologies, making them difficult to implement on a large scale. Webcams can be used as a substitution but are reported to have lower accuracy compared to precise eye trackers~\cite{eyetrackerWebcam}. Also, eye gaze contains rich personal information that the crowd workers might resist sharing~\cite{eyetrackerPrivacy}.
\par
An alternative method for capturing human attention involves human-in-the-loop systems, integrating human input directly into the model training process. For example, a user-interactive system employing scribble interactions in the medical imaging domain has been proposed to direct the model's focus to crucial image regions~\cite{scribble}. Similarly, another system allows users to interact with and efficiently redirect the attention of biased DNN models with minimal input~\cite{guide}. However, these approaches can be quite time-consuming, as they reported fine-tuning each model for several hours.
\par
Other techniques focus on using model attention to identify key areas within images. Initial research~\cite{ModelAttention1} and its subsequent extension~\cite{ModelAttention2} involved using the model's attention to pinpoint essential image areas. Additionally, a method was developed to divide images into patches and analyze model predictions on each to extract significant areas for classification~\cite{ModelAttention3}. These studies, however, solely rely on the trained classifier to determine important areas, not incorporating human attention, which could potentially increase reliability.
\par
Lastly, manual methods like drawing polygons on images have achieved permanent annotations of human attention. One study demonstrated the effectiveness of polygon annotations in guiding model attention in image segmentation tasks~\cite{segmentation}. Its extended work showed similar results using bounding box annotations~\cite{segmentation2}. However, such manual annotation methods are time-intensive and prone to quality issues, especially when implemented on a large scale~\cite{mturk}.
\par
Improving the works about human and model attention, we propose a patch-based labeling technique to extract the critical area from the image without auxiliary devices. Also, with AI assistance, we can quickly obtain more reliable human attention, as our experiments prove (\autoref{Experiments}).

\subsection{AI in Data Labeling}
To alleviate the cognitive and physical burden on human labelers, various AI-assisted systems for image class labeling have been developed. One approach involves synthesizing class-labeled images using basic features~\cite{SUFT}. However, it's noted that such labeled data often includes a high number of false positives requiring further post-processing~\cite{LabelGeneration}. Another strategy utilizes active learning techniques, where an image classifier, after initial human input, suggests labels for multiple samples~\cite{LabelBatch}. Additionally, trained classifiers have been employed to resolve discrepancies in data labeling outcomes from different participants~\cite{LabelConflictResolution}. However, these works only focused on quick labeling, whereas we aim to extract human attention through the labeling process.
\par
In the realm of object segmentation, AI models have been applied in various specialized domains, such as medical imaging for tasks like ophthalmic image segmentation~\cite{SegmentationEye} and radiotherapy~\cite{SegmentationRadio}. There's also been progress in training Generative Adversarial Networks to create new object segmentation datasets~\cite{SegmentationGAN}. Recent models have been designed to receive additional inputs like granularity or context in various forms. For example, a model that can adjust the granularity of segmentation tasks through a controllable parameter~\cite{SegmentationAnything}, and a multi-modality segmentation model that processes contextual information in different formats like text or scribbles~\cite{SegmentationSEEM}. However, these methods cannot fully replace human labeling due to potential inaccuracies in unique domains. Additionally, these models often struggle to achieve the desired granularity in images not used in their pre-training~\cite{celeba, awa2}, as described in~\autoref{fig:segmentAnything}.
\par
Despite numerous efforts to employ AI for data labeling and object segmentation, challenges persist due to the inherent uncertainties of AI systems~\cite{DeeplearningUnreliable} and misalignment to capture human-focused data. Consequently, this research incorporates a saliency detection model before the human labeling process to extract human attention while reducing time expenditure.

\section{Attention Extraction} \label{Attention Extraction}
We describe the detailed process of our human attention extraction algorithm. The pipeline comprises two subprocesses: \textit{Saliency Segmentation} and \textit{Attention Crowdsourcing}.

\subsection{Saliency Segmentation} \label{SaliencySegmentation}
Before extracting attention areas from crowd workers, we can reduce their labor by pre-attentively identifying the target object's location within the images. To achieve this, we performed a process called \textit{Saliency Segmentation}. In this step, we use a pre-trained salient object detection model to generate the initial saliency map. Then, we group adjacent pixels from the map to approximate the object's location.

\subsubsection{Salient Area Detection} \label{salient_area_detection}
\begin{figure}[t]
  \centering
  \includegraphics[width=0.7\linewidth]{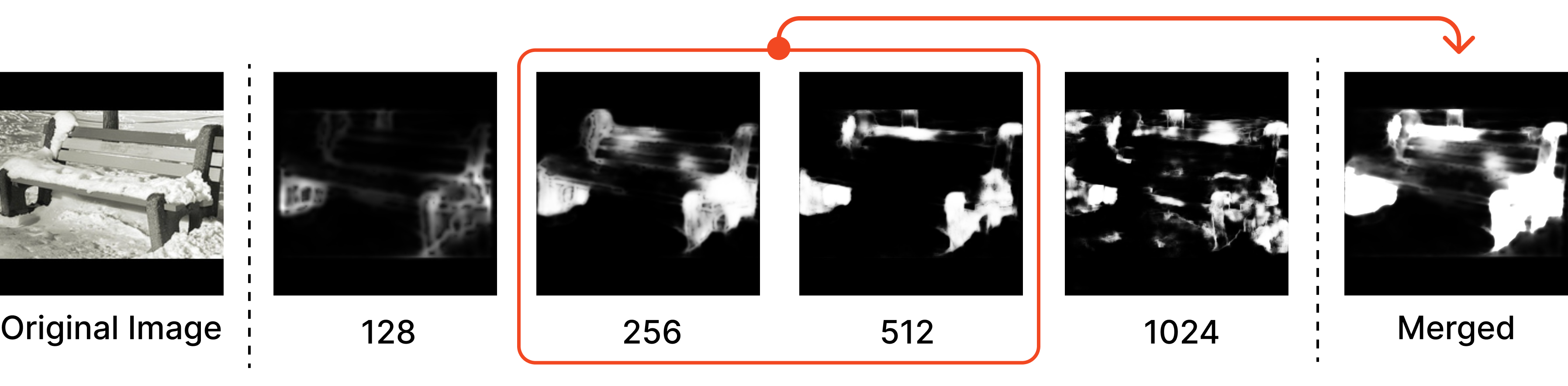}
  \caption{Generated saliency maps according to each image size, which differs in the quality and dispersion of pieces. We merged two image sizes, 256 and 512, to generate a saliency map covering a broad range and significant sections. Other sizes (128 or 1024) are not used because they produce blank or too scattered maps.}
  \Description{Generated saliency maps according to each image size, which differs in the quality and dispersion of the map. We merged two image sizes, 256 and 512, to generate a saliency map covering a broad range and significant sections. Other sizes (128 or 1024) are not used because they produce blank or too scattered maps.}
  \label{fig:saliency}
  \vspace{-5mm}
\end{figure}

We have tested various models for detecting a salient area from the image. Our choice was the Feature Pyramid-based model~\cite{SaliencyModern} due to its outstanding performance in generating clear saliency maps when intertwined objects appear. Earlier models produced unsatisfactory saliency maps, often failing to pinpoint the target object \cite{SaliencyPrev1} or not preserving its boundary shape \cite{SaliencyPrev2}. Also, we chose to use the mixture of saliency maps~(\autoref{fig:saliency}) generated from image sizes 256 and 512 to encompass both high and low-level information. This is based on our empirical findings that the model generates fine-grained but scattered maps for high-resolution images and broader but smoother maps for low-resolution images.
\par
After the saliency map is created, we verify whether the salient area of the image includes the given object by asking the crowd workers. If the model correctly predicted the salient area, they would successfully identify the object. However, if people cannot recognize the object, we can assume that the model had predicted a wrong area as salient or the image is in low resolution. In such cases, we skip the \textit{Pixel Clustering} step (\autoref{AttentionExtraction:PixelClustering}) and move on to the \textit{Attention Crowdsourcing} step (\autoref{AttentionExtraction:Attention Crowdsourcing}) and apply the overlap gridding on the entire image to find the target image.

\subsubsection{Pixel Clustering} \label{AttentionExtraction:PixelClustering}
As depicted in~\autoref{alg:pixel_clustering}, in this step, we group adjacent pixels and separate objects for the saliency maps reported to include the object. The pixels are grouped based on whether they are adjacent without blank spaces, following the Connected Component Labeling algorithm \cite{CCL}. Other various clustering algorithms were tested for our method, such as DBSCAN \cite{DBSCAN}, $K$-Means \cite{Kmeans}, and mean-shifting \cite{MeanShift}. However, despite their ability to cluster free-formed objects quickly, there were cases where two or more distinct objects were clustered together even though they were apart. 
\par
Pixel groups with larger sizes than \textit{minGroupSize}, which is set by the hyperparameter, are filtered and then distributed to crowd workers to verify whether each group includes the given object. According to their answers, we can filter out groups that do not include the target object, finally extracting the candidate area. However, if all pixel groups are labeled not to include the given object, we can conclude that all groups are necessary for the valid classification. Then, we skip the \textit{Attention Crowdsourcing} step and terminate the extraction process.

{\scriptsize
\begin{algorithm}
    \KwData{saliencyMap}
    \KwResult{targetSaliencyMap}
    \SetKwInOut{Hyperparameter}{Hyperparameter}
    \Hyperparameter{minGroupSize: minimum pixel group size}

    objects = ConnectedComponentLabeling(saliencyMap)\;
    targetSaliencyMap = \textit{Blank saliency map}\;
    \For{object in objects}{
        \If{object.size > minGroupSize}{
            responses = Crowdsource(object);
            \If{responses.any()}{
                targetSaliencyMap.add(object)\;
            }
        }
    }

    \If{targetSaliencyMap \text{is empty}}{
        \text{Terminate extraction process}
    }
    \Else{
        \text{Move on to Attention Crowdsourcing with} targetSaliencyMap
    }

    \caption{Pixel Clustering}
    \label{alg:pixel_clustering}
\end{algorithm}
}

\subsection{Attention Crowdsourcing} \label{AttentionExtraction:Attention Crowdsourcing}
After the \textit{Saliency Segmentation} step, we split images into small overlapped patches and repeatedly asked the crowd whether each patch contained the essential information for proper classification. In this step, \textit{Overlap Gridding}, \textit{Response Aggregation}, and \textit{Mask Update} are repetitively applied with shrinking grid size until it reaches the predefined minimum threshold. A full procedure of this step is described in \autoref{alg:attention}.
\subsubsection{Overlap Gridding}
Since previous steps tried to find where the object will likely be located automatically, \textit{Overlap Gridding} further narrows down the area. However, for cases where the \textit{Pixel Clustering} step is skipped due to the poorly generated saliency map, we consider the entire image salient, where all parts of the image become equally important.
\par
The image is divided into small patches using partially overlapped grids. Overlapping some portions of the patches ensures that the given object is not placed only in grid borders, preventing mislabeling. However, densely overlapping grids create many patches, leading to high labeling costs. Therefore, given the target patch size, we produced a minimum number of patches following~\autoref{eq:overlap_grid}.
{\scriptsize
    \begin{align}
        GridRows &= \lceil Height_{Image}/Height_{Patch} \rceil \nonumber \\
        GridCols &= \lceil Width_{Image}/Width_{Patch} \rceil \label{eq:overlap_grid}
    \end{align}
}
Also, when the saliency map generated from the previous steps exists, we calculate the IoU (Intersection over Union) score of the saliency map and each patch to filter out those with a score below the threshold. This ensures that all the produced patches will likely include the target object, leveraging the previous iteration's results. After properly filtering patches, we distribute the patches to the crowd and ask whether each patch contains the object at least partially. In this process, personal variance in perception might appear when determining whether each patch corresponds to the object. To manage this individual difference through collective intelligence, we question the same patch for more than one person and aggregate their responses.

\subsubsection{Response Aggregation} \label{Attention Extraction:Response Aggregation}
Hard or soft voting can be applied when aggregating the crowd's responses. A hard voting approach can create a discrete attention mask while losing the probabilistic information of how likely that patch is to contain the object. We, therefore, generate a continuous mask of human attention by adopting the soft voting algorithm in our approach, similar to the deep learning model's attention visualized with Class Activation Map (CAM)\cite{cam}.
\par
Moreover, we should measure the crowd workers' submission quality to filter out malicious users. We inject additional test images and score how accurately they classify them. If the validity score is reported below the predefined threshold, the responses from those workers are rejected and redeployed to the crowd to obtain an accurate attention mask. 

\subsubsection{Mask Update} \label{Attention Extraction:Mask Update}
By making the area bright for positive answers and dark for negative answers, we can generate a new response mask using the valid responses from the crowd workers. We then merge the attention mask generated in the previous iteration and the newly created one by applying two types of operation in~\autoref{eq:make_merge_full}: \textit{Weighted Summation} and \textit{Multiplication}. In the equation, $prev$ is the attention mask from the previous iteration, and $resp$ is the current response mask. \textit{Weighted Summation} preserves the information of the previous mask, which has a larger grid size, generating a smooth but wide mask. On the other hand, \textit{Multiplication} only leaves the overlapped area and creates a discrete but smaller mask. As we aim for our updated mask to be smooth to present the object's internal importance and discrete to contain its border information, we combined those operations for our final formula. It first divides the image into two sections where the previous mask is zero or not, and \textit{Weighted Summation} and \textit{Multiplication} are applied to each section. Also, to smoothen the vertical or horizontal lines created by grids, we applied Gaussian blur with the kernel size of the minimum grid size.

{\scriptsize
\begin{align}
    blurResp= GaussianBlur(resp) \nonumber \\
    updated_{i,j} =
    \begin{cases}
        (prev_{i,j} + blurResp_{i,j})/2 & \text{if } prev_{i,j} = 0 \\
        prev_{i,j} * blurResp_{i,j} & \text{if } prev_{i,j} \neq 0
    \end{cases} 
    \label{eq:make_merge_full}
\end{align}
}

{\scriptsize
\begin{algorithm}
    \KwData{image, targetSaliencyMap(from Algorithm 1)}
    \KwResult{attentionMask}
    \SetKwInOut{Hyperparameter}{Hyperparameter}
    \Hyperparameter{initSize: initial grid size, minSize: minimum grid size, shrinkRatio: shrink ratio, validThreshold: threshold for validity score, IoUThreshold: threshold for minimum IoU}

    objectSize = LargestObjectSize(targetSaliencyMap)\; 
    gridSize = min(initSize, objectSize)\; 
    previousMask = targetSaliencyMap\;
    
    \While{gridSize < minSize}{
        patches = OverlapGridding(image)\; 
        filteredPatches = FilterIoU(patches, previousMask, IoUThreshold)\;
        responses = CrowdSource(filteredPatches)\;
        responseMask = \textit{Mask created with responses.}\\
        validityScore = CheckValidity(responses)\; 
        \If {validityScore > validThreshold}{
            updatedMask = merge(previousMask, responseMask)\; 
            gridSize = gridSize * shrinkRatio\;
        }
        \Else {
            \textit{Reject answers and redeploy questions.}\\
            updatedMask = previousMask;\
        }
        previousMask = updatedMask
    }
    attentionMask = previousMask
    \caption{Attention Crowdsourcing}
    \label{alg:attention}
\end{algorithm}

}

\section{Experiments} \label{Experiments}
\subsection{Datasets} \label{Experiments:Datasets}
We tested our method to extract the human attention area on three major image datasets, COCO, AwA2, and CelebA, which were reported to include bias in their samples by prior works~\cite{COCOBias, AwA2Bias, CelebaBias}. The number of samples used for the attention extraction and model experiments is summarized in \autoref{tab:dataset}. It was relatively difficult to find samples that only included a single class object in the COCO dataset, so a small number of samples, which was 60, were used for the experiment. Detailed explanations about each dataset and how we sample biased data from it are described in the following sections.
{\scriptsize
\begin{table}[ht]
  \caption{Classes and number of samples selected from each dataset. Only biased samples are used for our experiment.}
  \label{tab:dataset}
  \begin{tabular}{c|l|c|c|c}
    \hline
    \textbf{Dataset} & \textbf{Classes} & \textbf{Train Split} & \textbf{Val Split} & \textbf{Test Split} \\
    \hline
    COCO & Boat, Horse & 60 & 60 & 60 \\
    CelebA & Wearing Lipstick, Mustache & 100 & 100 & 100 \\
    AwA2 & Horse, Humpback Whale & 100 & 100 & 100 \\
    \hline
  \end{tabular}
\end{table}
}

\subsubsection{COCO}
The Common Objects in Context (COCO) dataset \cite{coco} is constructed for object detection and semantic segmentation tasks. We selected the two target classes as \textit{Boat} or \textit{Horse}, which include the co-occurrence bias with their typical surroundings \cite{COCOBias}, such as water for boats and grass for horses. Due to its purpose of object detection, some of its images contain multiple classes, causing them to be intertwined. To properly evaluate our method, we only selected images that correspond to a single class, leaving the intended bias.

\subsubsection{CelebA}
The Large-scale CelebFaces Attributes (CelebA) dataset \cite{celeba} comprises facial images of celebrities and is utilized for classifying diverse facial traits. We chose the in-the-wild version without facial crop and alignment among various choices (e.g. CelebAMask-HQ \cite{celebaHQ}, Cropped \& Aligned \cite{celeba}). It often includes full-body images, which can be used to validate whether our method can effectively discern facial traits even from full-body poses. We selected \textit{Wearing Lipstick} and 
\textit{Mustache} from the many attributes available to leave only a reported bias factor: gender. Our exploration of data samples confirmed that all images labeled \textit{Wearing Lipstick} consisted of female figures, while images labeled \textit{Mustache} contained only males. 

\subsubsection{AwA2}
Animals With Attributes2 (AwA2) dataset \cite{awa2} consists of images of 80 animal species, each annotated with boolean features. We chose the \textit{Humpback Whale} and \textit{Horse} species as target classes due to reported biases. Most of the images of humpback whales are taken inside water, whereas many horse images feature humans alongside, indicating a co-occurrence bias. 

\subsection{Crowdsourcing} \label{Experiments:Crowdsourcing}
\begin{figure}[t]
  \centering
  \includegraphics[width=\linewidth]{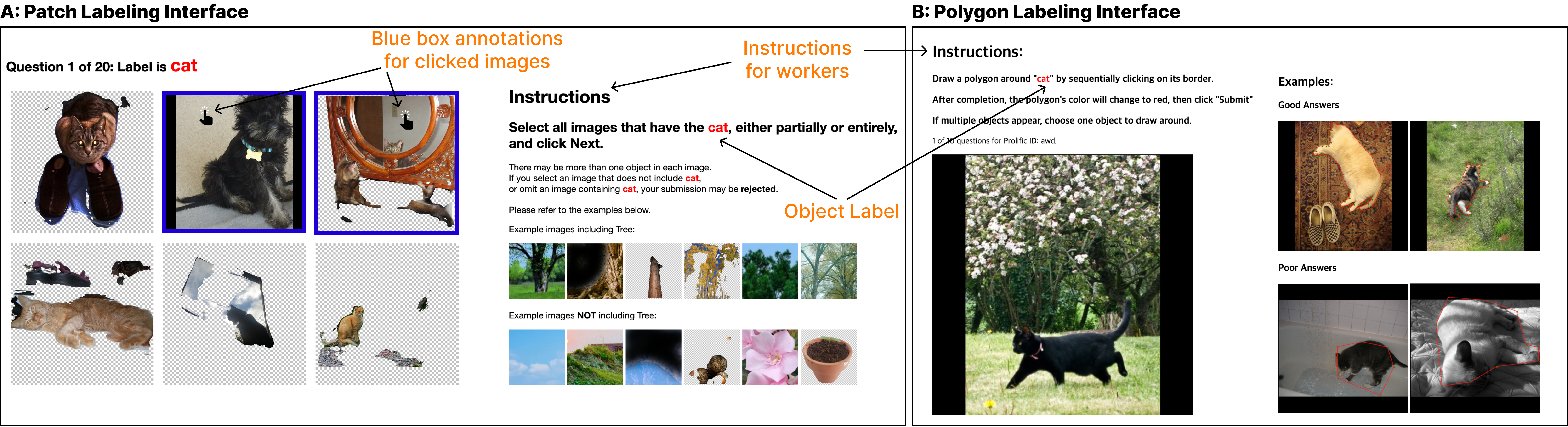}
  \caption{Prolific survey interface for our experiment containing questions shown to workers. Both the (A) patch labeling interface and (B) polygon drawing interface are described. Instructions are placed on the side screen for constant reminders. Object labels (Cat) appear in the instructions in red. The image is annotated with blue boxes if the user clicks on it, while users can freely draw a polygon by clicking the points.}
  \Description{}
  \label{fig:screen}

  \vspace{-3mm}
\end{figure}

\subsubsection{Platform \& Interface}
We used Prolific \cite{prolific} to deploy our experiments among the various crowdsourcing platforms, which was reported to provide high-quality data \cite{ProlificMturk}. To raise the quality of crowd submissions, we chose participants whose submission acceptance rates are higher than 95\%, with at least 500 submissions.
\par
We designed a simple labeling interface, shown in~\autoref{fig:screen}, for our experiments. Six images corresponding to the same class appear together on the question page to enhance the speed of the labeling process and reduce the workers' cognitive burden. Over six patches could not fit in a single screen, which required additional user scrolling interaction with increased labeling time. Each participant answered 20 questions pages to match the minimum task time set on the crowdsourcing platform~\cite{prolific}. The payment of the participants for the given labeling task is calculated as £6 per hour, based on the median time cost of the workers.

\subsubsection{Hyperparameters}
Hyperparameters listed in~\autoref{alg:pixel_clustering} and \autoref{alg:attention} must be set to run crowdsourced experiments. We conducted a pilot study to find adequate hyperparameters and set them as default for deployment in Prolific. To prevent bias from background knowledge, we tested a wide range of hyperparameters on seven graduate students majoring in computer science who were unfamiliar with the images from three datasets. We could derive useful, but not optimal, hyperparameters like the following. 
\par
The initial grid size was half the full image size ($initSize=128$). It is reduced at 40\% every iteration so that 25\% of the grid size can be overlapped ($shrinkRatio=0.4$). The grid size gradually shrinks to 80, which was reported to produce attention masks with just enough resolution to cover the object ($minSize=80$). The IoU threshold for filtering out the overlap grid patches was set to 0.3, based on the result of the pilot test that the participants could not distinguish the object if its size were smaller than 30\% of the patch size ($IoUThreshold=0.3$). The minimum group size for clustering pixels in~\autoref{AttentionExtraction:PixelClustering} was set as 32*32, which is about 1.5\% of the image size ($minGroupSize=1024$).
\par
Participants of our pilot study sometimes made mistakes when the images had low resolution or contained only part of the objects. Therefore, the validity score threshold for the submissions was set as 0.9, meaning the participants should label 9 test questions correctly when a total of 10 is given. We planned to reject the submissions with their validity scores below the threshold, allowing minor human errors during the labeling process. The effect of our rejection policy based on the validity scores is discussed in~\autoref{Discussion:Filtering}

\subsubsection{Time Cost Measurement}
We measured and aggregated the time required to label each patch on every step to assess the time cost for our crowdsourced human labeling. As six patches appear together on the screen for annotating the patches, the time cost for a single patch is calculated by dividing the page's time cost by six. We have averaged the results among participants to more precisely estimate the time cost for each patch, given that the patches were distributed randomly. Also, we analyze the per-class time cost to ensure that our result is not biased for several images.
\par
To prove our patch-based approach's efficiency in terms of time cost compared to object detection annotations, we have also deployed the object boundary annotation task (\autoref{fig:screen}B) to the Prolific and measured the time cost while labeling each image. To ensure that the drawn polygon has a sufficient shape that covers the object accurately, we have only accepted submissions with polygon annotations containing at least five key points. 

\subsubsection{Consensus Analysis}
We further analyze what causes the users' answers to differ in the crowdsourcing experiment. To evaluate how contentious the users' submitted answers are on the given patches, we calculate each patch's consensus score. It is calculated with the following~\autoref{eq:consensus}, giving a higher score to a patch without any controversy. $Ans$ represents the list of submitted answers on whether the patch contains the target object. The answer of \textit{true} corresponds to 1, and \textit{false} corresponds to -1.
{\scriptsize
\begin{align}
    ConsensusScore &= \frac{|\sum^{N}_{i=1}{Ans_i}|}{Ans.length}
    \label{eq:consensus}
\end{align}
}
A higher consensus score represents that the users tended to submit the same answers. Through measuring the consensus, we aim to discover which patches cause the users to be confused during \textit{Attention Crowdsourcing}. We also expect to verify the effect of soft voting, which smoothly incorporates the variance in perception.

\subsection{Bias Experiments with Image Classifiers}
Based on the extracted human attention mask from Prolific, we have tested the ability to fix the model bias problem by training the ResNet-50~\cite{resnet} model. We trained the network to classify each class pair in~\autoref{tab:dataset}.

\subsubsection{Hyperparameters}
We used the Adam optimizer with a learning rate of 0.001 for the COCO dataset and 0.0001 for the other two datasets. The difference is based on the sample size (\autoref{tab:dataset}) to prevent COCO from being overfitted. The model is trained for 50 epochs, and the checkpoint with the highest validation accuracy was reported for all attention guidance methods in the next section.

\subsubsection{Attention Guidance}
The model is trained with three methods: no guidance (Baseline), guidance with polygon (Polygon), and guidance with our attention mask (Attention) to check whether our method can solve the model bias problem. We only used the annotated images in~\autoref{Experiments:Crowdsourcing}, which was enough to train the model with sufficient performance and evaluate our method's effectiveness in guiding the model's attention. To guide the model using the extracted human attention, we first need to calculate which part of the image influences the model's prediction. We have adopted the CAM~\cite{cam} to derive which region the model focuses on in the image. Based on the calculated model attention, attention loss is designed to rectify the model's biased attention through a mean square error between the min-max scaled CAM and our extracted human attention mask. The final loss used for training the model is the weighted sum of cross entropy loss and the attention loss, performing both bias correction and proper classification. The detailed loss is described in \autoref{eq:full_loss} with the balance weight $w$=0.5, and the image width $W$ and height $H$. The balance weight is derived from He et al.~\cite{guide}, which equally mixed the attention guidance and cross-entropy loss.

{\scriptsize
\begin{align}
    &AttentionLoss = \frac{1}{WH}\sum_{i}^{W}\sum_{j}^{H} (CAM(model, img)_{i,j}-mask_{i,j})^2 \nonumber \\
    &FullLoss = w * CrossEntropy(model(img), label) + (1 - w) * AttentionLoss(img, mask) \label{eq:full_loss}
\end{align}
}

\begin{figure}[t]
  \centering
  \includegraphics[width=0.8\linewidth]{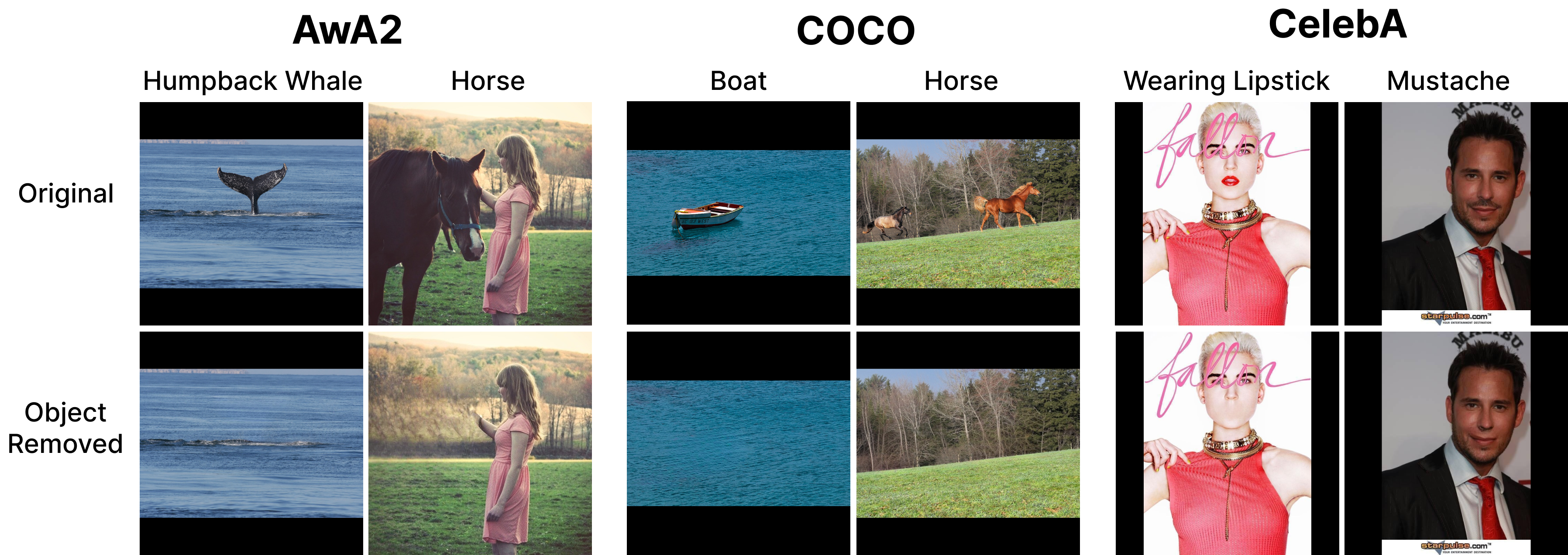}
  \caption{Target object is removed from the original image, only leaving the bias factors. Object auto removal service \cite{silverai} is used for the generation.}
  \Description{The first two columns are from the COCO dataset, with the horse and boat removed. The third and fourth columns are from the AwA2 dataset, with the horse and humpback whale removed. The last two columns are from the CelebA dataset with the mustache and lipstick removed.}
  \label{fig:background_images}

  \vspace{-3mm}
\end{figure}

\subsubsection{Bias Measurement}
 For the qualitative analysis of bias correction on two guidance methods (Polygon and Attention), we first visualized each model's attention as a heatmap using the CAM~\cite{cam}. We then analyzed which part of the image the model focuses on to make the class prediction. Also, to quantitatively verify whether the model's bias is removed, we have conducted two experiments. First, the average accuracy and F1-score are measured for three trials with random seeds and averaged to see whether the guided models have obtained the general knowledge of each class against the Baseline. Next, as Vaze et al. \cite{logit} empirically proved that the logit outputs can be a good estimator for out-of-domain images, we analyze the output of the model when the image, where the target class object is removed, is fed. Compared to the similar bias measurement by Rudd et al.~\cite{CelebALogit}, who masked the target object with a black rectangle, we used an AI-based object removal tool \cite{silverai} to remove the object from the image while preserving the other part of the image. \autoref{fig:background_images} shows the example images where the object is removed in each dataset. We have generated 20 samples for each class in all three datasets and measured their logits on Baseline, Polygon, and Attention.

\section{Results} \label{Results}
\subsection{Crowdsourcing}
\subsubsection{Human Attention Mask}

\begin{figure}[t]
  \centering
  \includegraphics[width=0.8\linewidth]{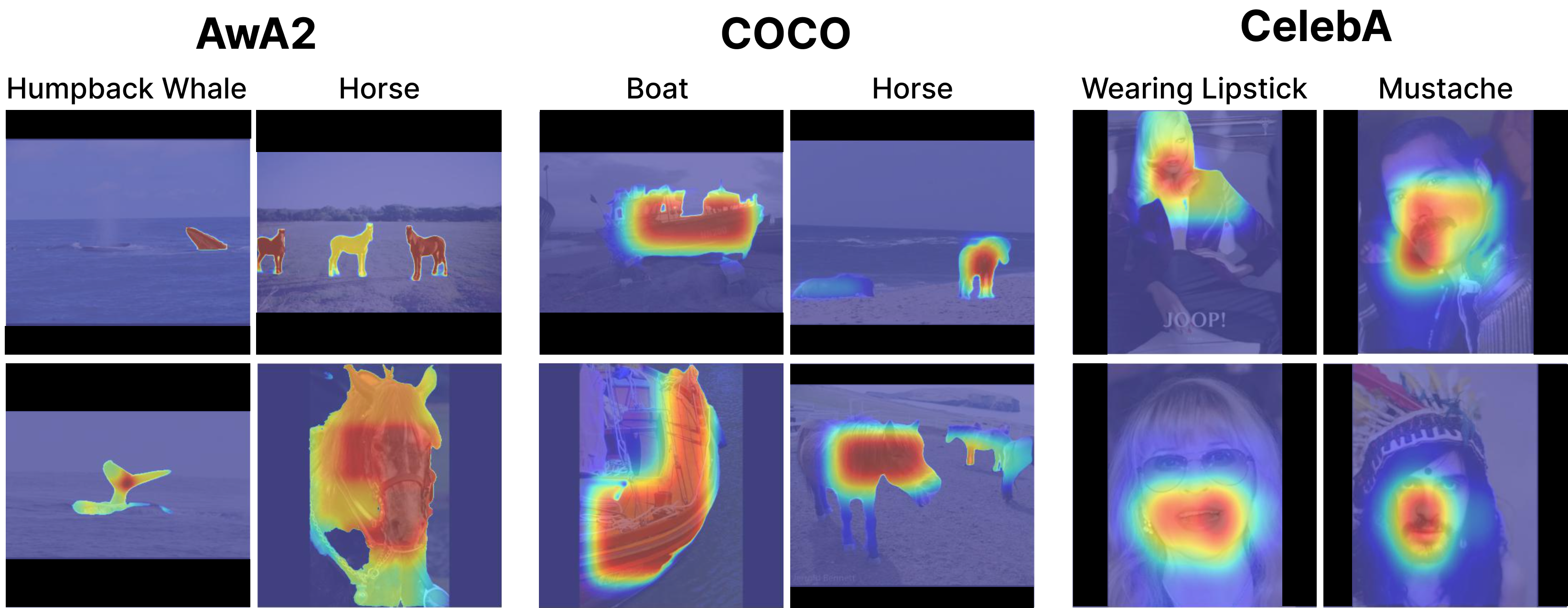}
  \caption{Examples of the extracted Human Attention masks on each dataset. The mask is overlapped with the original image as a heatmap, and red indicates having greater attention from humans.}
  \Description{AwA2 dataset: Humpback Whale (left) and Horse (right). COCO dataset: Boat (left) and Horse (right). CelebA dataset: Wearing Lipstick (left) and Mustache (right).}
  \label{fig:mask_samples}
\end{figure}

The retrieved human attention masks using our method are shown in \autoref{fig:mask_samples}. Due to the fine-grained saliency maps generated by the salient object detection model, the masks smoothly cover the object's border in the COCO and AwA2 datasets. Also, our method can capture multiple objects appearing, as seen in the third image of horses from the AwA2 dataset and the second image of horses from the COCO dataset. However, for the CelebA dataset, we have observed that the model predicts the entire face as a salient area in both classes. Therefore, the recurrent Overlap Gridding step further reduces the area to focus only on the lips and mustache.

\subsubsection{Time Cost Analysis} \label{Results:TimeCostAnalysis}

\begin{figure}[t]
  \centering
  \includegraphics[width=0.6\linewidth]{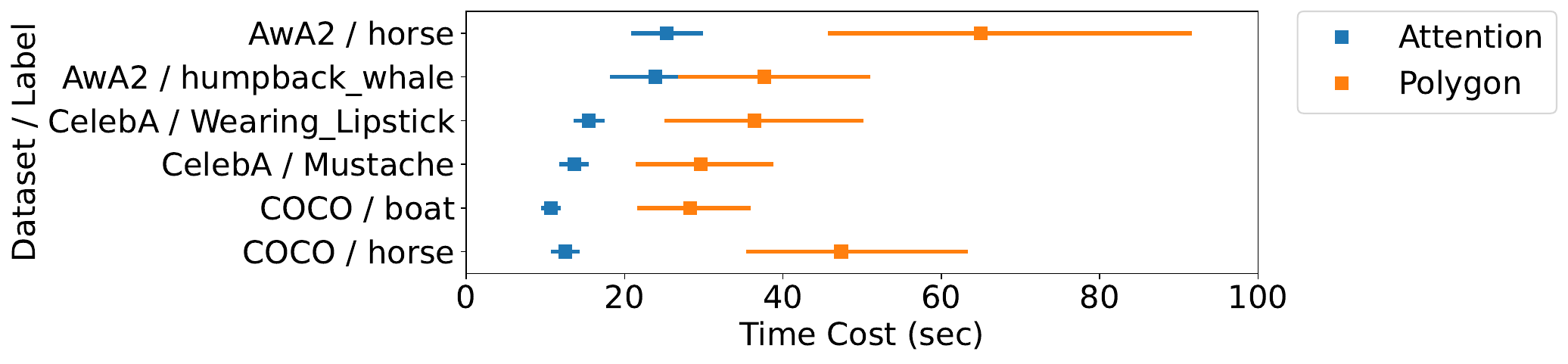}
  \caption{The time cost measured for obtaining each image annotation: Attention (Blue) and Polygon (Orange). The plotted error is within 95\% confidence.}
  \Description{}
  \label{fig:time_cost}

  \vspace{-3mm}
\end{figure}

 As shown in \autoref{fig:time_cost}, our method can significantly reduce the time spent on data annotation compared to drawing a polygon around the object, reducing overall human labor. Most crowdsourcing platforms, including Prolific, calculate the reward for the workers based on the expected completion time, so our method also has economic superiority compared to the polygon annotations. Also, we can further save the time cost by increasing the minimum grid size, having a trade-off between the mask resolution and time cost. We further discuss how to shorten the labeling time with parallelism in \autoref{Discussion:Parallelism}. Furthermore, our experiment indicates that while the time cost of our method remains consistent, the polygon method shows significant variability on all datasets. This suggests that polygon drawing is highly susceptible to individual differences, while our method can constantly show high efficiency even in the crowdsourcing environment. A comprehensive discussion on the personal variance regarding accessibility will be elaborated in section~\autoref{Discussion:Adaptivity}. Noticeably, the AwA2 dataset generally takes longer than the CelebA and COCO datasets. We concluded that the object size was larger on AwA2 datasets, taking longer due to the increased number of patches. However, small scattered objects usually appear in the COCO dataset for boat class, leading to fewer patches and time costs. We further discuss applying our method on various datasets with different object sizes in~\autoref{Discussion:Hyperparameter}.

\subsubsection{Consensus Analysis} \label{Results:Consensus}
\begin{figure}[t]
  \centering
  \includegraphics[width=0.8\linewidth]{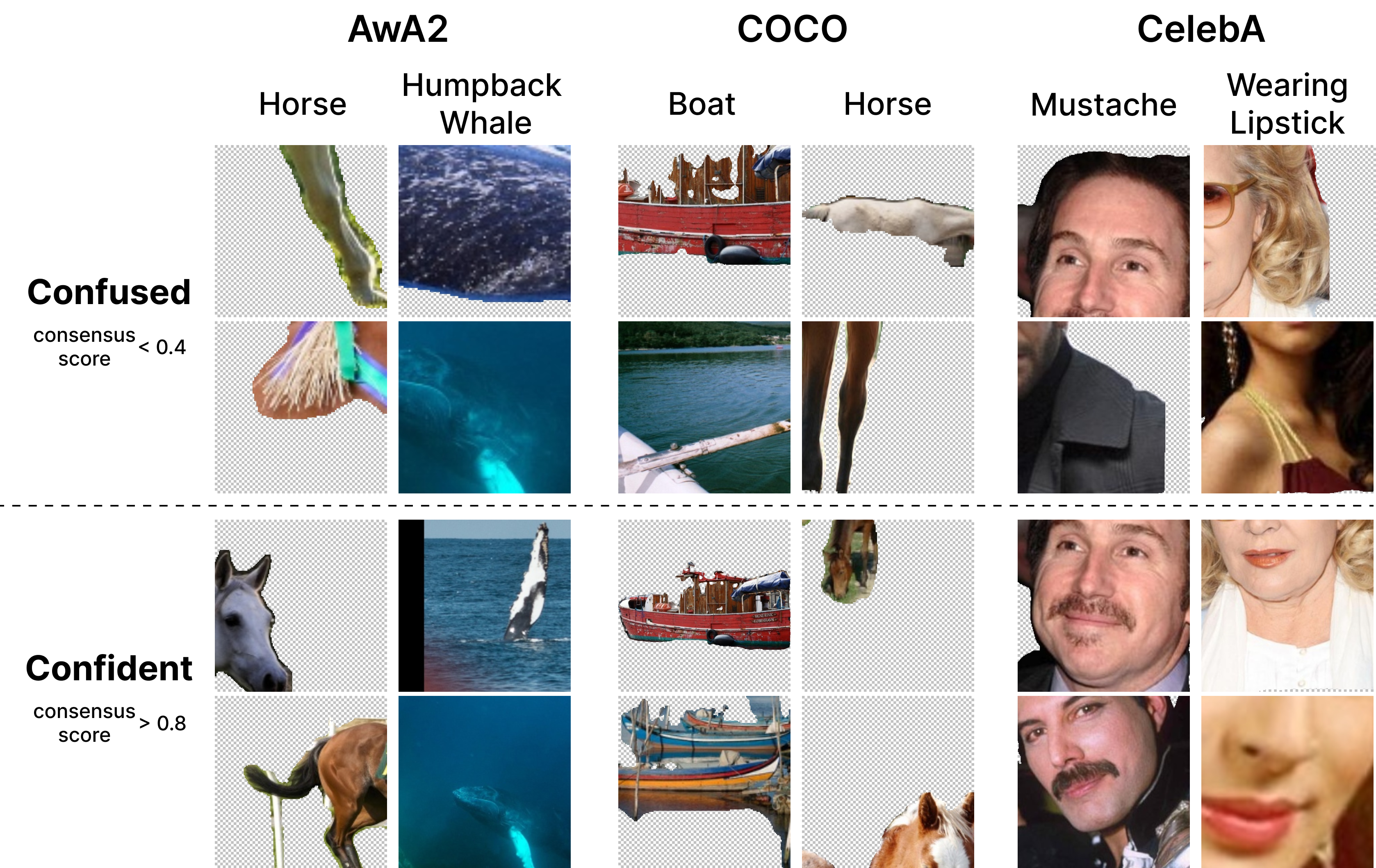}
  \caption{Sample of patches where users had low (<0.4) or high (>0.8) consensus score. The confusing patches contain only part of the target object or are in low resolution. Instead, confident patches fully contain the object or at least key features like a horse's pointy ears (3rd row, 4th column).}
  \Description{}
  
  \label{fig:consensus}
  \vspace{-3mm}
\end{figure}
We calculated consensus scores for all datasets and analyzed the patches that ranked high (over 0.8) or low (under 0.4) in~\autoref{fig:consensus}. We can observe that high-scored patches fully contain the target object or at least have the key feature that is used for people to identify. Compared, patches with low scores only partially include the target object or have a low resolution, generating a \textit{confusing region}. Soft voting in~\autoref{Attention Extraction:Mask Update} allows the final attention mask to incorporate variant perceptions of the crowd in such regions. The result is depicted in~\autoref{fig:mask_samples} as the yellow-colored area with intermediate importance.

\subsection{Bias Experiments with Image Classifiers}
\subsubsection{Classification Performance}
The metrics calculated are summarized in \autoref{tab:accuracy_f1}, showing 30 trials' average accuracy and F1-score. If our Attention approach guides the model's attention, the model accuracy becomes significantly higher than the Baseline for all three datasets. We applied the Mann‐Whitney U Test for statistical verification due to the non-normality. Its significance was proved on all datasets with $p=0.002$ (AwA2) and $p<0.001$ (COCO, CelebA). Surprisingly, there was a noticeable increase in the model performance for the CelebA (4.3\%p) and COCO (8.2\%p) datasets. These results show that the model can be robust when they are guided to focus on the same regions in the image that typically capture human attention.
{\scriptsize
\begin{table}[t]
  \centering
  \caption{Accuracy and F1 Score of our model experiment with each attention guidance method. A total of 30 trials are averaged.}
  \label{tab:accuracy_f1}
  \begin{tabular}{l|cccccc}
    \toprule
    \multirow{2}{*}{Method} & \multicolumn{2}{c}{AwA2} & \multicolumn{2}{c}{COCO} & \multicolumn{2}{c}{CelebA} \\
    \cmidrule(lr){2-3} \cmidrule(lr){4-5} \cmidrule(lr){6-7}
    & Accuracy & F1 Score & Accuracy & F1 Score & Accuracy & F1 Score \\
    \midrule 
    Baseline & 0.950 & 0.9499 & 0.680 & 0.6779 & 0.57 & 0.5696 \\
    Polygon & 0.9587 & 0.9587 & 0.7573 & 0.7549 & 0.6163 & 0.6067 \\
    Attention & 0.957 & 0.9570 & 0.762 & 0.7599 & 0.613 & 0.6061 \\
    \bottomrule
  \end{tabular}
\end{table}
}

\subsubsection{Model Attention Visualization}

\begin{figure}[t]
  \centering
  \includegraphics[width=\linewidth]{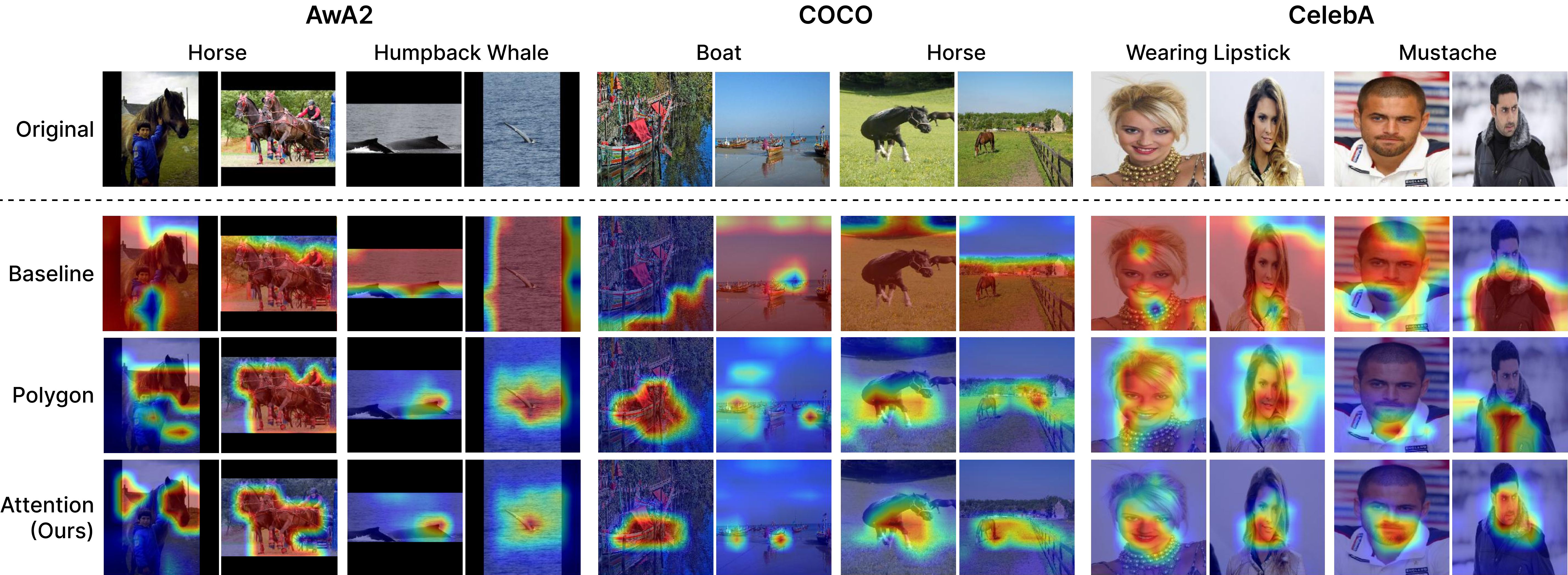}
  \caption{Visualized attention heatmap of models on images guided by each method. The red area, compared to the blue, has more attention from the trained model. The model guided with Attention shows a higher concentration on the object than Baseline and Polygon.}
  \Description{}
  \label{fig:guidance_samples}
\end{figure}

In all three datasets, the visualized CAM in \autoref{fig:guidance_samples} shows that the model bias is removed in both the Polygon and Attention approaches, having its attention concentrated mainly on the object. However, cases were observed where the Polygon approach did not completely solve the bias problem. For instance, in the CelebA dataset, the whole face was used to predict \textit{Wearing Lipstick}, while the male figure's jacket became a bias for \textit{Mustache}. Our method (Attention) successfully guided the model to focus on the target label in such cases. We will discuss this as the robustness in crowdsourcing in \autoref{Discussion:Filtering}.

\subsubsection{Measuring Bias Correction}

\begin{figure}[t]
  \centering
  \includegraphics[width=\linewidth]{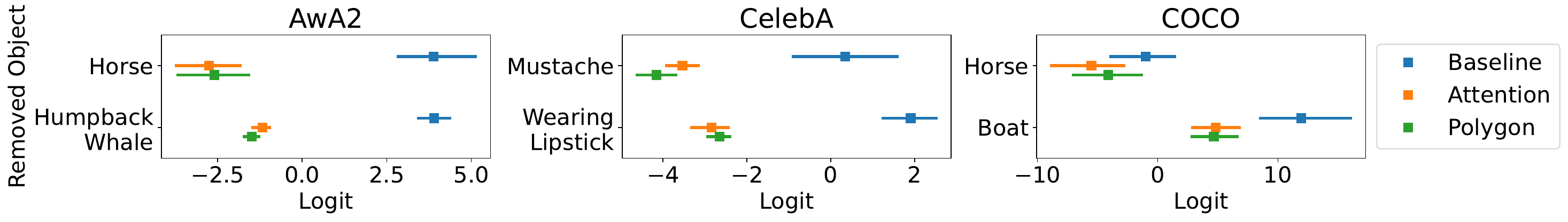}
  \caption{Logit values produced by models on each dataset only with bias factors. Models guided by the Polygon and Attention approach generate smaller logits, indicating they stay robust against bias.}
  \Description{}
  \label{fig:logit}

  \vspace{-3mm}
\end{figure}

The measured logits in \autoref{fig:logit} show that in most cases, the Polygon and Attention guidance force the model to predict less confidently when only the bias factor is shown. The model with Attention became less overconfident due to the additional attention loss used for the training. It performed bias correction and normalization effect, increasing the model's generalizability.

\section{Discussion} \label{Discussion}
We discuss our method's strengths within a crowdsourcing context, especially regarding parallelism and adaptive design. We also discuss the limitations of our research and possible enhancements.

\subsection{Maximizing Participant-Level Parallelism} \label{Discussion:Parallelism}
Data labeling can be performed much faster in large datasets by parallelizing the task to many workers. Due to Prolific's policy, we had to make the task longer than one minute to ensure proper compensation for the workers, resulting in 120 patch questions being allocated to each user. However, without this restriction, we could distribute the patches to more users, assigning a single patch to each individual. Compared to ours, the maximum parallelization achievable for object boundary annotations like polygons is at the image level. Based on the time cost calculated in \autoref{tab:scalability}, we can theoretically finish the dataset labeling up to 30 times faster than drawing polygons. This also allows the requester to distribute the labeling task among the workers in a more fine-grained and controlled manner, down to the minimum granularity of seconds. Increasing the level of parallelism can let the labeling tasks also be performed on daily smartphone devices as microtasks~\cite{microtask}. Further discussion of mobile crowdsourcing applications will be addressed in~\autoref{Discussion:Adaptivity}.

{\scriptsize
\begin{table}
  \caption{Theoretical time cost for labeling single image, when all methods are parallelized in maximum.}
  \label{tab:scalability}
  \begin{tabular}{c|c|c|c}
    \hline
    \textbf{Dataset} & \textbf{Attention (sec)} & \textbf{Polygon (sec)} & \textbf{Theoretical Speed Up} \\
    \hline
    COCO & 1.8429 & 39.6958 & x21\\
    CelebA & 1.5131 & 37.2271 & x25\\
    AwA2 & 1.5725 & 47.7901 & x30\\
    \hline
  \end{tabular}
  \vspace{-5mm}
\end{table}
}

\subsection{Advanced Salient Object Detection Models} \label{Discussion:SaliencyModel}
Among the various pre-trained saliency detection models, we selected the model that generated the cleanest and highest-quality saliency maps among those with official implementations and pre-trained weights. However, one of the drawbacks of these models is that they are usually trained with popular datasets that only include daily objects. The salient object detection model designed or fine-tuned for specific domains, such as medical \cite{SaliencyMedical} or robotics \cite{SaliencyRobotics}, can be used in the \textit{Saliency Segmentation} step of our method without changing the structure pipeline. In such cases, the time cost for labeling patches would be much more efficient because the produced saliency maps would be more accurate. 
\par
Moreover, if the state-of-the-art saliency detection model is used, we expect our method to be more time-efficient and adapted to various datasets. Advanced saliency detection models are being researched to generate saliency maps with more clear resolutions. Our approach can leverage such technical enhancements to further reduce human labor.

\subsection{Enhanced Accessibility and Adaptivity} \label{Discussion:Adaptivity}
Drawing annotations over the given image using digital devices can be difficult for people with accessibility issues. One study found that individuals with motor disabilities and older adults typically experience longer task completion times and higher error rates with tasks that involve pointing and dragging~\cite{Motor}. Also, our experiment results on~\autoref{fig:time_cost} indicate that personal variance in the completion time exists when users physically label (polygon drawing) the object. Instead, our method only requires a binary selection of the designated patch; thus, previously noted accessibility challenges are effectively alleviated.
\par
Moreover, our simple design allows robust adaptivity on various platforms, including mobile devices. These edge devices usually have a touchscreen interface, whose input can be inaccurate due to human motor constraints. This was proved by experiments by Yamanaka and Usuba \cite{TouchFittsLaw}, where a tradeoff between accuracy and speed was observed while performing repetitive touch-based tasks. Therefore, labeling based on polygons may exhibit vulnerability when executed on touchscreens, whereas our method maintains relative robustness. Additionally, our method can be readily extended to services such as CAPTCHA (Completely Automated Public Turing Test to Tell Computers and Humans Apart), often deployed on global devices. 
\par
Such adaptivity is also essential for maintaining a large pool of available workers in crowdsourcing. Having many potential participants is vital to finishing the labeling task quickly and having the collected answers less biased. Eye-tracking-based methods in \autoref{RelatedWork:HumanAttention} lose half of the participants because only 54.6\% participants in Prolific reported having a webcam on their device. Instead, our method can even enlarge the pool of participants to various touch-based devices as mobile crowdsourcing platforms and technologies are emerging~\cite{mobileCrowdsourcing, mobileCrowdsourcingPlatform}.

\subsection{Filtering Erroneous Submissions} \label{Discussion:Filtering}
\begin{figure}[t]
  \centering
  \includegraphics[width=0.7\linewidth]{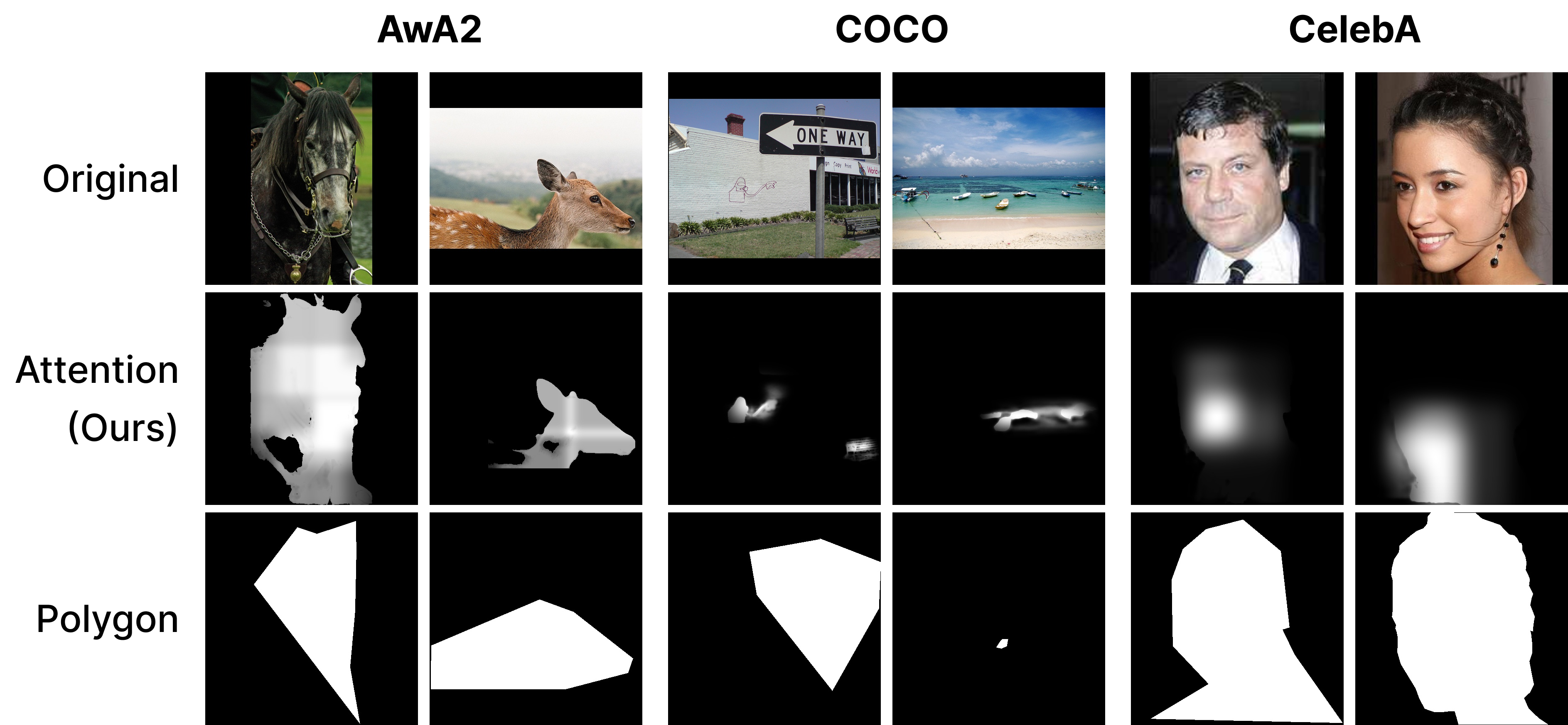}
  \caption{Cases where malicious users draw polygons poorly for various objects. The drawn polygons do not cover the intended object accurately compared to the extracted human attention mask.}
  \Description{}
  \label{fig:discussion}
  \vspace{-5mm}
\end{figure}

Our patch-labeling method in crowdsourcing efficiently manages low-quality submissions, including malicious users submitting minimal-effort answers and normal users making mistakes. To maintain data quality, it's essential to filter these submissions. While hand-picking answers are feasible for small volumes, an automatic filtering process is necessary for large datasets. This typically involves using dummy images with known labels for scoring. However, this approach is less practical for polygon drawing tasks, as judging the accuracy of the submitted vertex list is complex. During the crowdsourced experiment, we found that users often produce inaccurate polygons (\autoref{fig:discussion}). Additionally, filtering answers based on submission time is challenging due to the varying time taken by participants, especially when the object size is large. Thus, collecting multiple annotations per image becomes necessary, though it increases effort and cost.
\par
Our method effectively maintained high-quality data by automatically filtering out about 10\% of submissions with low validity scores during the experiment in Prolific. This process helps identify and remove randomly labeled patches by malicious users, underscoring the need for efficient filtering. Our approach also dilutes less accurate submissions by combining answers from multiple participants using soft voting (\autoref{Attention Extraction:Response Aggregation}), reducing the impact of poorer responses. Additionally, this method allows for evaluating submissions without dummy questions by assessing the alignment of participants' responses, helping to isolate outliers. Our future work could focus on quantitatively testing the robustness of this approach.

\subsection{Limitations and Future work}
Though we have planned our experiments based on thorough justification, we acknowledge that our study has several limitations and possible future works to address them. 

\subsubsection{Abstract Categories}
In our experiment, three datasets with clear object classes were used. However, there are cases where abstract labels need to be classified, such as emotions~\cite{emotionClassification} or artistic styles~\cite{artisticClassification}. The saliency detection model might not work correctly for cases requiring holistic insights. Also, the crowd worker's opinion about the target label may differ, making the generated attention mask ambiguous. Additional human efforts or sophisticated techniques that better capture human intention should be introduced to solve this problem.

\subsubsection{Usage for Data Augmentation}
We only used the extracted human attention area to fix the biased model. However, such information can be utilized in other tasks, such as data augmentation. For example, some of the previous image data augmentation techniques based on mixing image patches \cite{CutMix} can be improved using the obtained human attention information by only mixing the images' attentive area. Also, using the latest image generation AIs~\cite{kandinsky}, we can automatically generate new samples while preserving the important part of the image through inpainting~\cite{inpaint}. Expanding the usage of our method would be one of our future work.

\subsubsection{Optimal Hyperparameters} \label{Discussion:Hyperparameter}
Based on our pilot study, we defined the hyperparameters in~\autoref{Experiments:Crowdsourcing}. However, we have discovered in~\autoref{Results:TimeCostAnalysis} that some datasets with large objects generate massive amounts of patches, lowering the overall efficiency of our method. We can further increase our method's efficiency by adaptively changing the hyperparameters, such as the minimum grid size, according to the object size in the dataset. We can attentively estimate the object size of each class through the saliency model and adjust our hyperparameter to set the minimum patch size at the optimum level.

\section{Conclusion} \label{Conclusion}
In this study, we introduced a novel method for extracting the human attention area from images using a collaborative effort among humans with AI support. The core of our approach relies on a combination of saliency detection and crowdsourced patch-labeling, reducing the overall human effort. By employing this method, we overcome the biases found in many datasets, especially when used to train the DNN classifier model. Our experiments on three prominent image datasets demonstrated that guided models using our human attention can help achieve more accurate and less biased model performance. We also compared our approach with traditional polygon annotations and found our method more cost-efficient, particularly with crowdsourcing. We expect our method to be expanded to various platforms, users, and usages. In essence, our research highlights the importance and effectiveness of incorporating human insights into AI systems, leading to more balanced, fair, and efficient models.

\begin{acks}

\end{acks}

\bibliographystyle{ACM-Reference-Format}
\bibliography{_base}


\begin{thebibliography}{65}


\ifx \showCODEN    \undefined \def \showCODEN     #1{\unskip}     \fi
\ifx \showDOI      \undefined \def \showDOI       #1{#1}\fi
\ifx \showISBNx    \undefined \def \showISBNx     #1{\unskip}     \fi
\ifx \showISBNxiii \undefined \def \showISBNxiii  #1{\unskip}     \fi
\ifx \showISSN     \undefined \def \showISSN      #1{\unskip}     \fi
\ifx \showLCCN     \undefined \def \showLCCN      #1{\unskip}     \fi
\ifx \shownote     \undefined \def \shownote      #1{#1}          \fi
\ifx \showarticletitle \undefined \def \showarticletitle #1{#1}   \fi
\ifx \showURL      \undefined \def \showURL       {\relax}        \fi
\providecommand\bibfield[2]{#2}
\providecommand\bibinfo[2]{#2}
\providecommand\natexlab[1]{#1}
\providecommand\showeprint[2][]{arXiv:#2}

\bibitem[Alvi et~al\mbox{.}(2018)]%
        {BiasFeaturerepresentation}
\bibfield{author}{\bibinfo{person}{Mohsan Alvi}, \bibinfo{person}{Andrew Zisserman}, {and} \bibinfo{person}{Christoffer Nell{\aa}ker}.} \bibinfo{year}{2018}\natexlab{}.
\newblock \showarticletitle{Turning a blind eye: Explicit removal of biases and variation from deep neural network embeddings}. In \bibinfo{booktitle}{\emph{Proceedings of the European Conference on Computer Vision (ECCV) Workshops}}. \bibinfo{pages}{0--0}.
\newblock


\bibitem[Ashktorab et~al\mbox{.}(2021)]%
        {LabelBatch}
\bibfield{author}{\bibinfo{person}{Zahra Ashktorab}, \bibinfo{person}{Michael Desmond}, \bibinfo{person}{Josh Andres}, \bibinfo{person}{Michael Muller}, \bibinfo{person}{Narendra~Nath Joshi}, \bibinfo{person}{Michelle Brachman}, \bibinfo{person}{Aabhas Sharma}, \bibinfo{person}{Kristina Brimijoin}, \bibinfo{person}{Qian Pan}, \bibinfo{person}{Christine~T. Wolf}, \bibinfo{person}{Evelyn Duesterwald}, \bibinfo{person}{Casey Dugan}, \bibinfo{person}{Werner Geyer}, {and} \bibinfo{person}{Darrell Reimer}.} \bibinfo{year}{2021}\natexlab{}.
\newblock \showarticletitle{AI-Assisted Human Labeling: Batching for Efficiency without Overreliance}.
\newblock \bibinfo{journal}{\emph{Proc. ACM Hum.-Comput. Interact.}} \bibinfo{volume}{5}, \bibinfo{number}{CSCW1}, Article \bibinfo{articleno}{89} (\bibinfo{date}{apr} \bibinfo{year}{2021}), \bibinfo{numpages}{27}~pages.
\newblock
\urldef\tempurl%
\url{https://doi.org/10.1145/3449163}
\showDOI{\tempurl}


\bibitem[Bay et~al\mbox{.}(2006)]%
        {SUFT}
\bibfield{author}{\bibinfo{person}{Herbert Bay}, \bibinfo{person}{Tinne Tuytelaars}, {and} \bibinfo{person}{Luc Van~Gool}.} \bibinfo{year}{2006}\natexlab{}.
\newblock \showarticletitle{Surf: Speeded up robust features}. In \bibinfo{booktitle}{\emph{Computer Vision--ECCV 2006: 9th European Conference on Computer Vision, Graz, Austria, May 7-13, 2006. Proceedings, Part I 9}}. Springer, \bibinfo{pages}{404--417}.
\newblock


\bibitem[Brachman et~al\mbox{.}(2022)]%
        {LabelConflictResolution}
\bibfield{author}{\bibinfo{person}{Michelle Brachman}, \bibinfo{person}{Zahra Ashktorab}, \bibinfo{person}{Michael Desmond}, \bibinfo{person}{Evelyn Duesterwald}, \bibinfo{person}{Casey Dugan}, \bibinfo{person}{Narendra~Nath Joshi}, \bibinfo{person}{Qian Pan}, {and} \bibinfo{person}{Aabhas Sharma}.} \bibinfo{year}{2022}\natexlab{}.
\newblock \showarticletitle{Reliance and Automation for Human-AI Collaborative Data Labeling Conflict Resolution}.
\newblock \bibinfo{journal}{\emph{Proc. ACM Hum.-Comput. Interact.}} \bibinfo{volume}{6}, \bibinfo{number}{CSCW2}, Article \bibinfo{articleno}{321} (\bibinfo{date}{nov} \bibinfo{year}{2022}), \bibinfo{numpages}{27}~pages.
\newblock
\urldef\tempurl%
\url{https://doi.org/10.1145/3555212}
\showDOI{\tempurl}


\bibitem[Cardenas et~al\mbox{.}(2019)]%
        {SegmentationRadio}
\bibfield{author}{\bibinfo{person}{Carlos~E. Cardenas}, \bibinfo{person}{Jinzhong Yang}, \bibinfo{person}{Brian~M. Anderson}, \bibinfo{person}{Laurence~E. Court}, {and} \bibinfo{person}{Kristy~B. Brock}.} \bibinfo{year}{2019}\natexlab{}.
\newblock \showarticletitle{Advances in Auto-Segmentation}.
\newblock \bibinfo{journal}{\emph{Seminars in Radiation Oncology}} \bibinfo{volume}{29}, \bibinfo{number}{3} (\bibinfo{year}{2019}), \bibinfo{pages}{185--197}.
\newblock
\showISSN{1053-4296}
\urldef\tempurl%
\url{https://doi.org/10.1016/j.semradonc.2019.02.001}
\showDOI{\tempurl}
\newblock
\shownote{Adaptive Radiotherapy and Automation}.


\bibitem[Cetinic et~al\mbox{.}(2018)]%
        {artisticClassification}
\bibfield{author}{\bibinfo{person}{Eva Cetinic}, \bibinfo{person}{Tomislav Lipic}, {and} \bibinfo{person}{Sonja Grgic}.} \bibinfo{year}{2018}\natexlab{}.
\newblock \showarticletitle{Fine-tuning convolutional neural networks for fine art classification}.
\newblock \bibinfo{journal}{\emph{Expert Systems with Applications}}  \bibinfo{volume}{114} (\bibinfo{year}{2018}), \bibinfo{pages}{107--118}.
\newblock


\bibitem[Chi et~al\mbox{.}(2018)]%
        {mobileCrowdsourcing}
\bibfield{author}{\bibinfo{person}{Pei-Yu~(Peggy) Chi}, \bibinfo{person}{Anurag Batra}, {and} \bibinfo{person}{Maxwell Hsu}.} \bibinfo{year}{2018}\natexlab{}.
\newblock \showarticletitle{Mobile Crowdsourcing in the Wild: Challenges from a Global Community}. In \bibinfo{booktitle}{\emph{Proceedings of the 20th International Conference on Human-Computer Interaction with Mobile Devices and Services Adjunct}} (Barcelona, Spain) \emph{(\bibinfo{series}{MobileHCI '18})}. \bibinfo{publisher}{Association for Computing Machinery}, \bibinfo{address}{New York, NY, USA}, \bibinfo{pages}{410–415}.
\newblock
\showISBNx{9781450359412}
\urldef\tempurl%
\url{https://doi.org/10.1145/3236112.3236176}
\showDOI{\tempurl}


\bibitem[Comaniciu and Meer(2002)]%
        {MeanShift}
\bibfield{author}{\bibinfo{person}{D. Comaniciu} {and} \bibinfo{person}{P. Meer}.} \bibinfo{year}{2002}\natexlab{}.
\newblock \showarticletitle{Mean shift: a robust approach toward feature space analysis}.
\newblock \bibinfo{journal}{\emph{IEEE Transactions on Pattern Analysis and Machine Intelligence}} \bibinfo{volume}{24}, \bibinfo{number}{5} (\bibinfo{year}{2002}), \bibinfo{pages}{603--619}.
\newblock
\urldef\tempurl%
\url{https://doi.org/10.1109/34.1000236}
\showDOI{\tempurl}


\bibitem[Cornia et~al\mbox{.}(2018)]%
        {SaliencyPrev2}
\bibfield{author}{\bibinfo{person}{Marcella Cornia}, \bibinfo{person}{Lorenzo Baraldi}, \bibinfo{person}{Giuseppe Serra}, {and} \bibinfo{person}{Rita Cucchiara}.} \bibinfo{year}{2018}\natexlab{}.
\newblock \showarticletitle{Predicting Human Eye Fixations via an LSTM-Based Saliency Attentive Model}.
\newblock \bibinfo{journal}{\emph{IEEE Transactions on Image Processing}} \bibinfo{volume}{27}, \bibinfo{number}{10} (\bibinfo{year}{2018}), \bibinfo{pages}{5142--5154}.
\newblock
\urldef\tempurl%
\url{https://doi.org/10.1109/TIP.2018.2851672}
\showDOI{\tempurl}


\bibitem[Ester et~al\mbox{.}(1996)]%
        {DBSCAN}
\bibfield{author}{\bibinfo{person}{Martin Ester}, \bibinfo{person}{Hans-Peter Kriegel}, \bibinfo{person}{J\"{o}rg Sander}, {and} \bibinfo{person}{Xiaowei Xu}.} \bibinfo{year}{1996}\natexlab{}.
\newblock \showarticletitle{A Density-Based Algorithm for Discovering Clusters in Large Spatial Databases with Noise} \emph{(\bibinfo{series}{KDD'96})}. \bibinfo{publisher}{AAAI Press}, \bibinfo{pages}{226–231}.
\newblock


\bibitem[Eyal et~al\mbox{.}(2021)]%
        {ProlificMturk}
\bibfield{author}{\bibinfo{person}{Peer Eyal}, \bibinfo{person}{Rothschild David}, \bibinfo{person}{Gordon Andrew}, \bibinfo{person}{Evernden Zak}, {and} \bibinfo{person}{Damer Ekaterina}.} \bibinfo{year}{2021}\natexlab{}.
\newblock \showarticletitle{Data quality of platforms and panels for online behavioral research}.
\newblock \bibinfo{journal}{\emph{Behavior Research Methods}} (\bibinfo{year}{2021}), \bibinfo{pages}{1--20}.
\newblock


\bibitem[Findlater and Zhang(2020)]%
        {Motor}
\bibfield{author}{\bibinfo{person}{Leah Findlater} {and} \bibinfo{person}{Lotus Zhang}.} \bibinfo{year}{2020}\natexlab{}.
\newblock \showarticletitle{Input Accessibility: A Large Dataset and Summary Analysis of Age, Motor Ability and Input Performance}. In \bibinfo{booktitle}{\emph{Proceedings of the 22nd International ACM SIGACCESS Conference on Computers and Accessibility}} (Virtual Event, Greece) \emph{(\bibinfo{series}{ASSETS '20})}. \bibinfo{publisher}{Association for Computing Machinery}, \bibinfo{address}{New York, NY, USA}, Article \bibinfo{articleno}{17}, \bibinfo{numpages}{6}~pages.
\newblock
\showISBNx{9781450371032}
\urldef\tempurl%
\url{https://doi.org/10.1145/3373625.3417031}
\showDOI{\tempurl}


\bibitem[Gong et~al\mbox{.}(2012)]%
        {BiasUnsupervised}
\bibfield{author}{\bibinfo{person}{Boqing Gong}, \bibinfo{person}{Fei Sha}, {and} \bibinfo{person}{Kristen Grauman}.} \bibinfo{year}{2012}\natexlab{}.
\newblock \showarticletitle{Overcoming dataset bias: An unsupervised domain adaptation approach}. In \bibinfo{booktitle}{\emph{NIPS Workshop on Large Scale Visual Recognition and Retrieval}}, Vol.~\bibinfo{volume}{3}. Citeseer.
\newblock


\bibitem[He et~al\mbox{.}(2016)]%
        {resnet}
\bibfield{author}{\bibinfo{person}{Kaiming He}, \bibinfo{person}{Xiangyu Zhang}, \bibinfo{person}{Shaoqing Ren}, {and} \bibinfo{person}{Jian Sun}.} \bibinfo{year}{2016}\natexlab{}.
\newblock \showarticletitle{Deep residual learning for image recognition}. In \bibinfo{booktitle}{\emph{Proceedings of the IEEE conference on computer vision and pattern recognition}}. \bibinfo{pages}{770--778}.
\newblock


\bibitem[He et~al\mbox{.}(2023)]%
        {guide}
\bibfield{author}{\bibinfo{person}{Yi He}, \bibinfo{person}{Xi Yang}, \bibinfo{person}{Chia-Ming Chang}, \bibinfo{person}{Haoran Xie}, {and} \bibinfo{person}{Takeo Igarashi}.} \bibinfo{year}{2023}\natexlab{}.
\newblock \showarticletitle{Efficient Human-in-the-Loop System for Guiding DNNs Attention}. In \bibinfo{booktitle}{\emph{Proceedings of the 28th International Conference on Intelligent User Interfaces}} (Sydney, NSW, Australia) \emph{(\bibinfo{series}{IUI '23})}. \bibinfo{publisher}{Association for Computing Machinery}, \bibinfo{address}{New York, NY, USA}, \bibinfo{pages}{294–306}.
\newblock
\showISBNx{9798400701061}
\urldef\tempurl%
\url{https://doi.org/10.1145/3581641.3584074}
\showDOI{\tempurl}


\bibitem[Karras et~al\mbox{.}(2017)]%
        {celebaHQ}
\bibfield{author}{\bibinfo{person}{Tero Karras}, \bibinfo{person}{Timo Aila}, \bibinfo{person}{Samuli Laine}, {and} \bibinfo{person}{Jaakko Lehtinen}.} \bibinfo{year}{2017}\natexlab{}.
\newblock \showarticletitle{Progressive growing of gans for improved quality, stability, and variation}.
\newblock \bibinfo{journal}{\emph{arXiv preprint arXiv:1710.10196}} (\bibinfo{year}{2017}).
\newblock


\bibitem[Khosla et~al\mbox{.}(2012)]%
        {BiasUnrolling}
\bibfield{author}{\bibinfo{person}{Aditya Khosla}, \bibinfo{person}{Tinghui Zhou}, \bibinfo{person}{Tomasz Malisiewicz}, \bibinfo{person}{Alexei~A Efros}, {and} \bibinfo{person}{Antonio Torralba}.} \bibinfo{year}{2012}\natexlab{}.
\newblock \showarticletitle{Undoing the damage of dataset bias}. In \bibinfo{booktitle}{\emph{Computer Vision--ECCV 2012: 12th European Conference on Computer Vision, Florence, Italy, October 7-13, 2012, Proceedings, Part I 12}}. Springer, \bibinfo{pages}{158--171}.
\newblock


\bibitem[Kim et~al\mbox{.}(2019)]%
        {BiasNormalization}
\bibfield{author}{\bibinfo{person}{Byungju Kim}, \bibinfo{person}{Hyunwoo Kim}, \bibinfo{person}{Kyungsu Kim}, \bibinfo{person}{Sungjin Kim}, {and} \bibinfo{person}{Junmo Kim}.} \bibinfo{year}{2019}\natexlab{}.
\newblock \showarticletitle{Learning Not to Learn: Training Deep Neural Networks With Biased Data}. In \bibinfo{booktitle}{\emph{2019 IEEE/CVF Conference on Computer Vision and Pattern Recognition (CVPR)}}. \bibinfo{pages}{9004--9012}.
\newblock
\urldef\tempurl%
\url{https://doi.org/10.1109/CVPR.2019.00922}
\showDOI{\tempurl}


\bibitem[Kirillov et~al\mbox{.}(2023)]%
        {SegmentationAnything}
\bibfield{author}{\bibinfo{person}{Alexander Kirillov}, \bibinfo{person}{Eric Mintun}, \bibinfo{person}{Nikhila Ravi}, \bibinfo{person}{Hanzi Mao}, \bibinfo{person}{Chloe Rolland}, \bibinfo{person}{Laura Gustafson}, \bibinfo{person}{Tete Xiao}, \bibinfo{person}{Spencer Whitehead}, \bibinfo{person}{Alexander~C. Berg}, \bibinfo{person}{Wan-Yen Lo}, \bibinfo{person}{Piotr Dollár}, {and} \bibinfo{person}{Ross Girshick}.} \bibinfo{year}{2023}\natexlab{}.
\newblock \bibinfo{title}{Segment Anything}.
\newblock
\newblock
\showeprint[arxiv]{2304.02643}~[cs.CV]


\bibitem[Kr{\"o}ger et~al\mbox{.}(2020)]%
        {eyetrackerPrivacy}
\bibfield{author}{\bibinfo{person}{Jacob~Leon Kr{\"o}ger}, \bibinfo{person}{Otto Hans-Martin Lutz}, {and} \bibinfo{person}{Florian M{\"u}ller}.} \bibinfo{year}{2020}\natexlab{}.
\newblock \showarticletitle{What does your gaze reveal about you? On the privacy implications of eye tracking}.
\newblock \bibinfo{journal}{\emph{Privacy and Identity Management. Data for Better Living: AI and Privacy: 14th IFIP WG 9.2, 9.6/11.7, 11.6/SIG 9.2. 2 International Summer School, Windisch, Switzerland, August 19--23, 2019, Revised Selected Papers 14}} (\bibinfo{year}{2020}), \bibinfo{pages}{226--241}.
\newblock


\bibitem[Lang et~al\mbox{.}(2012)]%
        {SaliencyPrev1}
\bibfield{author}{\bibinfo{person}{Congyan Lang}, \bibinfo{person}{Tam~V. Nguyen}, \bibinfo{person}{Harish Katti}, \bibinfo{person}{Karthik Yadati}, \bibinfo{person}{Mohan Kankanhalli}, {and} \bibinfo{person}{Shuicheng Yan}.} \bibinfo{year}{2012}\natexlab{}.
\newblock \showarticletitle{Depth Matters: Influence of Depth Cues on Visual Saliency}. In \bibinfo{booktitle}{\emph{Computer Vision -- ECCV 2012}}, \bibfield{editor}{\bibinfo{person}{Andrew Fitzgibbon}, \bibinfo{person}{Svetlana Lazebnik}, \bibinfo{person}{Pietro Perona}, \bibinfo{person}{Yoichi Sato}, {and} \bibinfo{person}{Cordelia Schmid}} (Eds.). \bibinfo{publisher}{Springer Berlin Heidelberg}, \bibinfo{address}{Berlin, Heidelberg}, \bibinfo{pages}{101--115}.
\newblock


\bibitem[Li and Wang(2023)]%
        {ModelAttention2}
\bibfield{author}{\bibinfo{person}{Feiyang Li} {and} \bibinfo{person}{Jiangtao Wang}.} \bibinfo{year}{2023}\natexlab{}.
\newblock \showarticletitle{Remote Sensing Image Scene Classification via Regional Growth-Based Key Area Fine Location and Multilayer Feature Fusion}.
\newblock \bibinfo{journal}{\emph{IEEE Geoscience and Remote Sensing Letters}}  \bibinfo{volume}{20} (\bibinfo{year}{2023}), \bibinfo{pages}{1--5}.
\newblock
\urldef\tempurl%
\url{https://doi.org/10.1109/LGRS.2022.3233374}
\showDOI{\tempurl}


\bibitem[Li et~al\mbox{.}(2022)]%
        {ModelAttention1}
\bibfield{author}{\bibinfo{person}{Feiyang Li}, \bibinfo{person}{Jiangtao Wang}, \bibinfo{person}{Mingyang Wang}, \bibinfo{person}{Ziyang Wang}, {and} \bibinfo{person}{Mohammad~R. Khosravi}.} \bibinfo{year}{2022}\natexlab{}.
\newblock \showarticletitle{Single-Object-Based Region Growth: Key Area Localization Model for Remote Sensing Image Scene Classification}.
\newblock   \bibinfo{volume}{2022} (\bibinfo{date}{jan} \bibinfo{year}{2022}), \bibinfo{numpages}{9}~pages.
\newblock
\showISSN{1687-5680}
\urldef\tempurl%
\url{https://doi.org/10.1155/2022/5816565}
\showDOI{\tempurl}


\bibitem[Li et~al\mbox{.}(2018)]%
        {segmentation}
\bibfield{author}{\bibinfo{person}{Kunpeng Li}, \bibinfo{person}{Ziyan Wu}, \bibinfo{person}{Kuan-Chuan Peng}, \bibinfo{person}{Jan Ernst}, {and} \bibinfo{person}{Yun Fu}.} \bibinfo{year}{2018}\natexlab{}.
\newblock \showarticletitle{Tell me where to look: Guided attention inference network}. In \bibinfo{booktitle}{\emph{Proceedings of the IEEE conference on computer vision and pattern recognition}}. \bibinfo{pages}{9215--9223}.
\newblock


\bibitem[Li et~al\mbox{.}(2019)]%
        {segmentation2}
\bibfield{author}{\bibinfo{person}{Kunpeng Li}, \bibinfo{person}{Ziyan Wu}, \bibinfo{person}{Kuan-Chuan Peng}, \bibinfo{person}{Jan Ernst}, {and} \bibinfo{person}{Yun Fu}.} \bibinfo{year}{2019}\natexlab{}.
\newblock \showarticletitle{Guided attention inference network}.
\newblock \bibinfo{journal}{\emph{IEEE transactions on pattern analysis and machine intelligence}} \bibinfo{volume}{42}, \bibinfo{number}{12} (\bibinfo{year}{2019}), \bibinfo{pages}{2996--3010}.
\newblock


\bibitem[Li et~al\mbox{.}(2023)]%
        {multipleShortcuts}
\bibfield{author}{\bibinfo{person}{Zhiheng Li}, \bibinfo{person}{Ivan Evtimov}, \bibinfo{person}{Albert Gordo}, \bibinfo{person}{Caner Hazirbas}, \bibinfo{person}{Tal Hassner}, \bibinfo{person}{Cristian~Canton Ferrer}, \bibinfo{person}{Chenliang Xu}, {and} \bibinfo{person}{Mark Ibrahim}.} \bibinfo{year}{2023}\natexlab{}.
\newblock \showarticletitle{A whac-a-mole dilemma: Shortcuts come in multiples where mitigating one amplifies others}. In \bibinfo{booktitle}{\emph{Proceedings of the IEEE/CVF Conference on Computer Vision and Pattern Recognition}}. \bibinfo{pages}{20071--20082}.
\newblock


\bibitem[Lin et~al\mbox{.}(2014)]%
        {coco}
\bibfield{author}{\bibinfo{person}{Tsung-Yi Lin}, \bibinfo{person}{Michael Maire}, \bibinfo{person}{Serge Belongie}, \bibinfo{person}{James Hays}, \bibinfo{person}{Pietro Perona}, \bibinfo{person}{Deva Ramanan}, \bibinfo{person}{Piotr Doll{\'a}r}, {and} \bibinfo{person}{C.~Lawrence Zitnick}.} \bibinfo{year}{2014}\natexlab{}.
\newblock \showarticletitle{Microsoft COCO: Common Objects in Context}. In \bibinfo{booktitle}{\emph{Computer Vision -- ECCV 2014}}, \bibfield{editor}{\bibinfo{person}{David Fleet}, \bibinfo{person}{Tomas Pajdla}, \bibinfo{person}{Bernt Schiele}, {and} \bibinfo{person}{Tinne Tuytelaars}} (Eds.). \bibinfo{publisher}{Springer International Publishing}, \bibinfo{address}{Cham}, \bibinfo{pages}{740--755}.
\newblock
\showISBNx{978-3-319-10602-1}


\bibitem[Liu et~al\mbox{.}(2022)]%
        {EyeTrackingHierarchy}
\bibfield{author}{\bibinfo{person}{Yang Liu}, \bibinfo{person}{Lei Zhou}, \bibinfo{person}{Pengcheng Zhang}, \bibinfo{person}{Xiao Bai}, \bibinfo{person}{Lin Gu}, \bibinfo{person}{Xiaohan Yu}, \bibinfo{person}{Jun Zhou}, {and} \bibinfo{person}{Edwin~R. Hancock}.} \bibinfo{year}{2022}\natexlab{}.
\newblock \showarticletitle{Where To Focus: Investigating Hierarchical Attention Relationship For Fine-Grained Visual Classification}. In \bibinfo{booktitle}{\emph{Computer Vision – ECCV 2022: 17th European Conference, Tel Aviv, Israel, October 23–27, 2022, Proceedings, Part XXIV}} (Tel Aviv, Israel). \bibinfo{publisher}{Springer-Verlag}, \bibinfo{address}{Berlin, Heidelberg}, \bibinfo{pages}{57–73}.
\newblock
\showISBNx{978-3-031-20052-6}
\urldef\tempurl%
\url{https://doi.org/10.1007/978-3-031-20053-3_4}
\showDOI{\tempurl}


\bibitem[Liu et~al\mbox{.}(2015)]%
        {celeba}
\bibfield{author}{\bibinfo{person}{Ziwei Liu}, \bibinfo{person}{Ping Luo}, \bibinfo{person}{Xiaogang Wang}, {and} \bibinfo{person}{Xiaoou Tang}.} \bibinfo{year}{2015}\natexlab{}.
\newblock \showarticletitle{Deep Learning Face Attributes in the Wild}. In \bibinfo{booktitle}{\emph{Proceedings of International Conference on Computer Vision (ICCV)}}.
\newblock


\bibitem[Lloyd(1982)]%
        {Kmeans}
\bibfield{author}{\bibinfo{person}{S. Lloyd}.} \bibinfo{year}{1982}\natexlab{}.
\newblock \showarticletitle{Least squares quantization in PCM}.
\newblock \bibinfo{journal}{\emph{IEEE Transactions on Information Theory}} \bibinfo{volume}{28}, \bibinfo{number}{2} (\bibinfo{year}{1982}), \bibinfo{pages}{129--137}.
\newblock
\urldef\tempurl%
\url{https://doi.org/10.1109/TIT.1982.1056489}
\showDOI{\tempurl}


\bibitem[Long et~al\mbox{.}(2015)]%
        {BiasAdaptation}
\bibfield{author}{\bibinfo{person}{Mingsheng Long}, \bibinfo{person}{Yue Cao}, \bibinfo{person}{Jianmin Wang}, {and} \bibinfo{person}{Michael Jordan}.} \bibinfo{year}{2015}\natexlab{}.
\newblock \showarticletitle{Learning transferable features with deep adaptation networks}. In \bibinfo{booktitle}{\emph{International conference on machine learning}}. PMLR, \bibinfo{pages}{97--105}.
\newblock


\bibitem[Maji(2011)]%
        {mturk}
\bibfield{author}{\bibinfo{person}{Subhransu Maji}.} \bibinfo{year}{2011}\natexlab{}.
\newblock \showarticletitle{Large scale image annotations on amazon mechanical turk}.
\newblock \bibinfo{journal}{\emph{EECS Department, University of California, Berkeley, Tech. Rep. UCB/EECS-2011-79}} (\bibinfo{year}{2011}).
\newblock


\bibitem[Manerikar and Kak(2023)]%
        {SegmentationGAN}
\bibfield{author}{\bibinfo{person}{Ankit Manerikar} {and} \bibinfo{person}{Avinash~C Kak}.} \bibinfo{year}{2023}\natexlab{}.
\newblock \showarticletitle{Self-Supervised One-Shot Learning for Automatic Segmentation of StyleGAN Images}.
\newblock \bibinfo{journal}{\emph{arXiv preprint arXiv:2303.05639}} (\bibinfo{year}{2023}).
\newblock


\bibitem[M{\"u}ller et~al\mbox{.}(2023)]%
        {SaliencyVSModel}
\bibfield{author}{\bibinfo{person}{Romy M{\"u}ller}, \bibinfo{person}{Marcel Duerschmidt}, \bibinfo{person}{Julian Ullrich}, \bibinfo{person}{Carsten Knoll}, \bibinfo{person}{Sascha Weber}, {and} \bibinfo{person}{Steffen Seitz}.} \bibinfo{year}{2023}\natexlab{}.
\newblock \showarticletitle{Do humans and Convolutional Neural Networks attend to similar areas during scene classification: Effects of task and image type}.
\newblock \bibinfo{journal}{\emph{ArXiv}}  \bibinfo{volume}{abs/2307.13345} (\bibinfo{year}{2023}).
\newblock
\urldef\tempurl%
\url{https://api.semanticscholar.org/CorpusID:260155115}
\showURL{%
\tempurl}


\bibitem[Musthag and Ganesan(2013)]%
        {microtask}
\bibfield{author}{\bibinfo{person}{Mohamed Musthag} {and} \bibinfo{person}{Deepak Ganesan}.} \bibinfo{year}{2013}\natexlab{}.
\newblock \showarticletitle{Labor dynamics in a mobile micro-task market}. In \bibinfo{booktitle}{\emph{Proceedings of the SIGCHI conference on human factors in computing systems}}. \bibinfo{pages}{641--650}.
\newblock


\bibitem[Novozámský et~al\mbox{.}(2020)]%
        {LabelGeneration}
\bibfield{author}{\bibinfo{person}{A. Novozámský}, \bibinfo{person}{D. Vít}, \bibinfo{person}{F. Šroubek}, \bibinfo{person}{J. Franc}, \bibinfo{person}{M. Krbálek}, \bibinfo{person}{Z. Bílkova}, {and} \bibinfo{person}{B. Zitová}.} \bibinfo{year}{2020}\natexlab{}.
\newblock \showarticletitle{Automated Object Labeling For Cnn-Based Image Segmentation}. In \bibinfo{booktitle}{\emph{2020 IEEE International Conference on Image Processing (ICIP)}}. \bibinfo{pages}{2036--2040}.
\newblock
\urldef\tempurl%
\url{https://doi.org/10.1109/ICIP40778.2020.9191320}
\showDOI{\tempurl}


\bibitem[Prolific(2023)]%
        {prolific}
\bibfield{author}{\bibinfo{person}{Prolific}.} \bibinfo{year}{2023}\natexlab{}.
\newblock \bibinfo{booktitle}{\emph{Prolific: A Crowdsourcing Platform}}.
\newblock
\urldef\tempurl%
\url{https://www.prolific.com/}
\showURL{%
\tempurl}


\bibitem[Rajabi et~al\mbox{.}(2023)]%
        {BiasDebiasModel}
\bibfield{author}{\bibinfo{person}{Amirarsalan Rajabi}, \bibinfo{person}{Mehdi Yazdani-Jahromi}, \bibinfo{person}{Ozlem~Ozmen Garibay}, {and} \bibinfo{person}{Gita Sukthankar}.} \bibinfo{year}{2023}\natexlab{}.
\newblock \showarticletitle{Through a Fair Looking-Glass: Mitigating Bias in Image Datasets}. In \bibinfo{booktitle}{\emph{Artificial Intelligence in HCI}}, \bibfield{editor}{\bibinfo{person}{Helmut Degen} {and} \bibinfo{person}{Stavroula Ntoa}} (Eds.). \bibinfo{publisher}{Springer Nature Switzerland}, \bibinfo{address}{Cham}, \bibinfo{pages}{446--459}.
\newblock
\showISBNx{978-3-031-35891-3}


\bibitem[Razzhigaev et~al\mbox{.}(2023)]%
        {kandinsky}
\bibfield{author}{\bibinfo{person}{Anton Razzhigaev}, \bibinfo{person}{Arseniy Shakhmatov}, \bibinfo{person}{Anastasia Maltseva}, \bibinfo{person}{Vladimir Arkhipkin}, \bibinfo{person}{Igor Pavlov}, \bibinfo{person}{Ilya Ryabov}, \bibinfo{person}{Angelina Kuts}, \bibinfo{person}{Alexander Panchenko}, \bibinfo{person}{Andrey Kuznetsov}, {and} \bibinfo{person}{Denis Dimitrov}.} \bibinfo{year}{2023}\natexlab{}.
\newblock \bibinfo{title}{Kandinsky: an Improved Text-to-Image Synthesis with Image Prior and Latent Diffusion}.
\newblock
\newblock
\showeprint[arxiv]{2310.03502}~[cs.CV]


\bibitem[Rong et~al\mbox{.}(2021)]%
        {SaliencyClassification}
\bibfield{author}{\bibinfo{person}{Yao Rong}, \bibinfo{person}{Wenjia Xu}, \bibinfo{person}{Zeynep Akata}, {and} \bibinfo{person}{Enkelejda Kasneci}.} \bibinfo{year}{2021}\natexlab{}.
\newblock \showarticletitle{Human Attention in Fine-grained Classification}. In \bibinfo{booktitle}{\emph{British Machine Vision Conference}}.
\newblock
\urldef\tempurl%
\url{https://api.semanticscholar.org/CorpusID:240419768}
\showURL{%
\tempurl}


\bibitem[Rosenfeld and Pfaltz(1966)]%
        {CCL}
\bibfield{author}{\bibinfo{person}{Azriel Rosenfeld} {and} \bibinfo{person}{John~L Pfaltz}.} \bibinfo{year}{1966}\natexlab{}.
\newblock \showarticletitle{Sequential operations in digital picture processing}.
\newblock \bibinfo{journal}{\emph{Journal of the ACM (JACM)}} \bibinfo{volume}{13}, \bibinfo{number}{4} (\bibinfo{year}{1966}), \bibinfo{pages}{471--494}.
\newblock


\bibitem[Rudd et~al\mbox{.}(2016)]%
        {CelebALogit}
\bibfield{author}{\bibinfo{person}{Ethan Rudd}, \bibinfo{person}{Manuel Günther}, {and} \bibinfo{person}{Terrance Boult}.} \bibinfo{year}{2016}\natexlab{}.
\newblock \showarticletitle{MOON: A Mixed Objective Optimization Network for the Recognition of Facial Attributes}, Vol.~\bibinfo{volume}{9909}.
\newblock
\showISBNx{978-3-319-46453-4}
\urldef\tempurl%
\url{https://doi.org/10.1007/978-3-319-46454-1_2}
\showDOI{\tempurl}


\bibitem[Saxena et~al\mbox{.}(2023)]%
        {eyetrackerWebcam}
\bibfield{author}{\bibinfo{person}{Shreshth Saxena}, \bibinfo{person}{Lauren~K Fink}, {and} \bibinfo{person}{Elke~B Lange}.} \bibinfo{year}{2023}\natexlab{}.
\newblock \showarticletitle{Deep learning models for webcam eye tracking in online experiments}.
\newblock \bibinfo{journal}{\emph{Behavior Research Methods}} (\bibinfo{year}{2023}), \bibinfo{pages}{1--17}.
\newblock


\bibitem[Sharma et~al\mbox{.}(2022)]%
        {SegmentationEye}
\bibfield{author}{\bibinfo{person}{Parmanand Sharma}, \bibinfo{person}{Takahiro Ninomiya}, \bibinfo{person}{Kazuko Omodaka}, \bibinfo{person}{Naoki Takahashi}, \bibinfo{person}{Takehiro Miya}, \bibinfo{person}{Noriko Himori}, \bibinfo{person}{Takayuki Okatani}, {and} \bibinfo{person}{Toru Nakazawa}.} \bibinfo{year}{2022}\natexlab{}.
\newblock \showarticletitle{A lightweight deep learning model for automatic segmentation and analysis of ophthalmic images}.
\newblock \bibinfo{journal}{\emph{Scientific reports}} \bibinfo{volume}{12}, \bibinfo{number}{1} (\bibinfo{year}{2022}), \bibinfo{pages}{8508}.
\newblock


\bibitem[Shen et~al\mbox{.}(2021)]%
        {scribble}
\bibfield{author}{\bibinfo{person}{Haifeng Shen}, \bibinfo{person}{Kewen Liao}, \bibinfo{person}{Zhibin Liao}, \bibinfo{person}{Job Doornberg}, \bibinfo{person}{Maoying Qiao}, \bibinfo{person}{Anton van~den Hengel}, {and} \bibinfo{person}{Johan~W. Verjans}.} \bibinfo{year}{2021}\natexlab{}.
\newblock \showarticletitle{Human-AI Interactive and Continuous Sensemaking: A Case Study of Image Classification Using Scribble Attention Maps}. In \bibinfo{booktitle}{\emph{Extended Abstracts of the 2021 CHI Conference on Human Factors in Computing Systems}} (Yokohama, Japan) \emph{(\bibinfo{series}{CHI EA '21})}. \bibinfo{publisher}{Association for Computing Machinery}, \bibinfo{address}{New York, NY, USA}, Article \bibinfo{articleno}{290}, \bibinfo{numpages}{8}~pages.
\newblock
\showISBNx{9781450380959}
\urldef\tempurl%
\url{https://doi.org/10.1145/3411763.3451798}
\showDOI{\tempurl}


\bibitem[SilverAI(2023)]%
        {silverai}
\bibfield{author}{\bibinfo{person}{SilverAI}.} \bibinfo{year}{2023}\natexlab{}.
\newblock \bibinfo{booktitle}{\emph{SnapEdit: Object Deletion}}.
\newblock
\urldef\tempurl%
\url{https://www.snapedit.app}
\showURL{%
\tempurl}


\bibitem[Sooch and Anand(2021)]%
        {emotionClassification}
\bibfield{author}{\bibinfo{person}{Shardeep~Kaur Sooch} {and} \bibinfo{person}{Darpan Anand}.} \bibinfo{year}{2021}\natexlab{}.
\newblock \showarticletitle{Emotion Classification and Facial Key point detection using AI}. In \bibinfo{booktitle}{\emph{2021 2nd International Conference on Advances in Computing, Communication, Embedded and Secure Systems (ACCESS)}}. \bibinfo{pages}{1--5}.
\newblock
\urldef\tempurl%
\url{https://doi.org/10.1109/ACCESS51619.2021.9563289}
\showDOI{\tempurl}


\bibitem[Torralba and Efros(2011)]%
        {COCOBias}
\bibfield{author}{\bibinfo{person}{Antonio Torralba} {and} \bibinfo{person}{Alexei~A. Efros}.} \bibinfo{year}{2011}\natexlab{}.
\newblock \showarticletitle{Unbiased look at dataset bias}. In \bibinfo{booktitle}{\emph{CVPR 2011}}. \bibinfo{pages}{1521--1528}.
\newblock
\urldef\tempurl%
\url{https://doi.org/10.1109/CVPR.2011.5995347}
\showDOI{\tempurl}


\bibitem[Vaze et~al\mbox{.}(2022)]%
        {logit}
\bibfield{author}{\bibinfo{person}{Sagar Vaze}, \bibinfo{person}{Kai Han}, \bibinfo{person}{Andrea Vedaldi}, {and} \bibinfo{person}{Andrew Zisserman}.} \bibinfo{year}{2022}\natexlab{}.
\newblock \showarticletitle{Open-Set Recognition: a Good Closed-Set Classifier is All You Need?}. In \bibinfo{booktitle}{\emph{International Conference on Learning Representations}}.
\newblock


\bibitem[Vermeire et~al\mbox{.}(2022)]%
        {ModelAttention3}
\bibfield{author}{\bibinfo{person}{Tom Vermeire}, \bibinfo{person}{Dieter Brughmans}, \bibinfo{person}{Sofie Goethals}, \bibinfo{person}{Raphael Mazzine~Barbossa de Oliveira}, {and} \bibinfo{person}{David Martens}.} \bibinfo{year}{2022}\natexlab{}.
\newblock \showarticletitle{Explainable image classification with evidence counterfactual}.
\newblock \bibinfo{journal}{\emph{Pattern Analysis and Applications}} \bibinfo{volume}{25}, \bibinfo{number}{2} (\bibinfo{year}{2022}), \bibinfo{pages}{315--335}.
\newblock


\bibitem[von Eschenbach(2021)]%
        {DeeplearningUnreliable}
\bibfield{author}{\bibinfo{person}{Warren~J von Eschenbach}.} \bibinfo{year}{2021}\natexlab{}.
\newblock \showarticletitle{Transparency and the black box problem: Why we do not trust AI}.
\newblock \bibinfo{journal}{\emph{Philosophy \& Technology}} \bibinfo{volume}{34}, \bibinfo{number}{4} (\bibinfo{year}{2021}), \bibinfo{pages}{1607--1622}.
\newblock


\bibitem[Wen et~al\mbox{.}(2023)]%
        {SaliencyRobotics}
\bibfield{author}{\bibinfo{person}{Falin Wen}, \bibinfo{person}{Qinghui Wang}, \bibinfo{person}{Ruirui Zou}, \bibinfo{person}{Ying Wang}, \bibinfo{person}{Fenglin Liu}, \bibinfo{person}{Yang Chen}, \bibinfo{person}{Linghao Yu}, \bibinfo{person}{Shaoyi Du}, {and} \bibinfo{person}{Chengzhi Yuan}.} \bibinfo{year}{2023}\natexlab{}.
\newblock \showarticletitle{A Salient Object Detection Method Based on Boundary Enhancement}.
\newblock \bibinfo{journal}{\emph{Sensors}} \bibinfo{volume}{23}, \bibinfo{number}{16} (\bibinfo{year}{2023}), \bibinfo{pages}{7077}.
\newblock


\bibitem[Xian et~al\mbox{.}(2019)]%
        {awa2}
\bibfield{author}{\bibinfo{person}{Yongqin Xian}, \bibinfo{person}{Christoph~H. Lampert}, \bibinfo{person}{Bernt Schiele}, {and} \bibinfo{person}{Zeynep Akata}.} \bibinfo{year}{2019}\natexlab{}.
\newblock \showarticletitle{Zero-Shot Learning—A Comprehensive Evaluation of the Good, the Bad and the Ugly}.
\newblock \bibinfo{journal}{\emph{IEEE Transactions on Pattern Analysis and Machine Intelligence}} \bibinfo{volume}{41}, \bibinfo{number}{9} (\bibinfo{year}{2019}), \bibinfo{pages}{2251--2265}.
\newblock
\urldef\tempurl%
\url{https://doi.org/10.1109/TPAMI.2018.2857768}
\showDOI{\tempurl}


\bibitem[Xu et~al\mbox{.}(2023)]%
        {SaliencyMedical}
\bibfield{author}{\bibinfo{person}{Cheng Xu}, \bibinfo{person}{Hui Wang}, \bibinfo{person}{Xianhui Liu}, {and} \bibinfo{person}{Weidong Zhao}.} \bibinfo{year}{2023}\natexlab{}.
\newblock \showarticletitle{Bi-attention network for bi-directional salient object detection}.
\newblock \bibinfo{journal}{\emph{Applied Intelligence}} (\bibinfo{year}{2023}), \bibinfo{pages}{1--17}.
\newblock


\bibitem[Yamanaka and Usuba(2022)]%
        {TouchFittsLaw}
\bibfield{author}{\bibinfo{person}{Shota Yamanaka} {and} \bibinfo{person}{Hiroki Usuba}.} \bibinfo{year}{2022}\natexlab{}.
\newblock \showarticletitle{Computing Touch-Point Ambiguity on Mobile Touchscreens for Modeling Target Selection Times}.
\newblock \bibinfo{journal}{\emph{Proc. ACM Interact. Mob. Wearable Ubiquitous Technol.}} \bibinfo{volume}{5}, \bibinfo{number}{4}, Article \bibinfo{articleno}{186} (\bibinfo{date}{dec} \bibinfo{year}{2022}), \bibinfo{numpages}{21}~pages.
\newblock
\urldef\tempurl%
\url{https://doi.org/10.1145/3494976}
\showDOI{\tempurl}


\bibitem[Yang et~al\mbox{.}(2022)]%
        {SaliencyImitator}
\bibfield{author}{\bibinfo{person}{Yi Yang}, \bibinfo{person}{Yueyuan Zheng}, \bibinfo{person}{Didan Deng}, \bibinfo{person}{Jindi Zhang}, \bibinfo{person}{Yongxiang Huang}, \bibinfo{person}{Yumeng Yang}, \bibinfo{person}{Janet~H. Hsiao}, {and} \bibinfo{person}{Caleb~Chen Cao}.} \bibinfo{year}{2022}\natexlab{}.
\newblock \showarticletitle{HSI: Human Saliency Imitator for Benchmarking Saliency-Based Model Explanations}.
\newblock \bibinfo{journal}{\emph{Proceedings of the AAAI Conference on Human Computation and Crowdsourcing}} \bibinfo{volume}{10}, \bibinfo{number}{1} (\bibinfo{date}{Oct.} \bibinfo{year}{2022}), \bibinfo{pages}{231--242}.
\newblock
\urldef\tempurl%
\url{https://doi.org/10.1609/hcomp.v10i1.22002}
\showDOI{\tempurl}


\bibitem[Ye et~al\mbox{.}(2017)]%
        {mobileCrowdsourcingPlatform}
\bibfield{author}{\bibinfo{person}{Kai Ye}, \bibinfo{person}{Yuling Sun}, \bibinfo{person}{Jing Yang}, {and} \bibinfo{person}{Liang He}.} \bibinfo{year}{2017}\natexlab{}.
\newblock \showarticletitle{WeCrowd: A WeChat based mobile crowdsourcing platform}. In \bibinfo{booktitle}{\emph{2017 IEEE 21st International Conference on Computer Supported Cooperative Work in Design (CSCWD)}}. \bibinfo{pages}{30--35}.
\newblock
\urldef\tempurl%
\url{https://doi.org/10.1109/CSCWD.2017.8066666}
\showDOI{\tempurl}


\bibitem[Yu et~al\mbox{.}(2023)]%
        {inpaint}
\bibfield{author}{\bibinfo{person}{Tao Yu}, \bibinfo{person}{Runseng Feng}, \bibinfo{person}{Ruoyu Feng}, \bibinfo{person}{Jinming Liu}, \bibinfo{person}{Xin Jin}, \bibinfo{person}{Wenjun Zeng}, {and} \bibinfo{person}{Zhibo Chen}.} \bibinfo{year}{2023}\natexlab{}.
\newblock \bibinfo{title}{Inpaint Anything: Segment Anything Meets Image Inpainting}.
\newblock
\newblock
\showeprint[arxiv]{2304.06790}~[cs.CV]


\bibitem[Yun et~al\mbox{.}(2019)]%
        {CutMix}
\bibfield{author}{\bibinfo{person}{Sangdoo Yun}, \bibinfo{person}{Dongyoon Han}, \bibinfo{person}{Seong~Joon Oh}, \bibinfo{person}{Sanghyuk Chun}, \bibinfo{person}{Junsuk Choe}, {and} \bibinfo{person}{Youngjoon Yoo}.} \bibinfo{year}{2019}\natexlab{}.
\newblock \showarticletitle{CutMix: Regularization Strategy to Train Strong Classifiers with Localizable Features}.
\newblock \bibinfo{journal}{\emph{CoRR}}  \bibinfo{volume}{abs/1905.04899} (\bibinfo{year}{2019}).
\newblock
\showeprint[arXiv]{1905.04899}
\urldef\tempurl%
\url{http://arxiv.org/abs/1905.04899}
\showURL{%
\tempurl}


\bibitem[Zhang et~al\mbox{.}(2017)]%
        {CelebaBias}
\bibfield{author}{\bibinfo{person}{Quanshi Zhang}, \bibinfo{person}{Wenguan Wang}, {and} \bibinfo{person}{Song-Chun Zhu}.} \bibinfo{year}{2017}\natexlab{}.
\newblock \showarticletitle{Examining CNN representations with respect to Dataset Bias}. In \bibinfo{booktitle}{\emph{AAAI Conference on Artificial Intelligence}}.
\newblock
\urldef\tempurl%
\url{https://api.semanticscholar.org/CorpusID:6347939}
\showURL{%
\tempurl}


\bibitem[Zhang et~al\mbox{.}(2020)]%
        {GazeSurvey}
\bibfield{author}{\bibinfo{person}{Ruohan Zhang}, \bibinfo{person}{Akanksha Saran}, \bibinfo{person}{Bo Liu}, \bibinfo{person}{Yifeng Zhu}, \bibinfo{person}{Sihang Guo}, \bibinfo{person}{Scott Niekum}, \bibinfo{person}{Dana Ballard}, {and} \bibinfo{person}{Mary Hayhoe}.} \bibinfo{year}{2020}\natexlab{}.
\newblock \showarticletitle{Human Gaze Assisted Artificial Intelligence: A Review}. In \bibinfo{booktitle}{\emph{Proceedings of the Twenty-Ninth International Joint Conference on Artificial Intelligence, {IJCAI-20}}}, \bibfield{editor}{\bibinfo{person}{Christian Bessiere}} (Ed.). \bibinfo{publisher}{International Joint Conferences on Artificial Intelligence Organization}, \bibinfo{pages}{4951--4958}.
\newblock
\urldef\tempurl%
\url{https://doi.org/10.24963/ijcai.2020/689}
\showDOI{\tempurl}
\newblock
\shownote{Survey track}.


\bibitem[Zhao et~al\mbox{.}(2018)]%
        {AwA2Bias}
\bibfield{author}{\bibinfo{person}{Bo Zhao}, \bibinfo{person}{Yanwei Fu}, \bibinfo{person}{Rui Liang}, \bibinfo{person}{Jiahong Wu}, \bibinfo{person}{Yonggang Wang}, {and} \bibinfo{person}{Yizhou Wang}.} \bibinfo{year}{2018}\natexlab{}.
\newblock \showarticletitle{A Large-Scale Attribute Dataset for Zero-Shot Learning}.
\newblock \bibinfo{journal}{\emph{2019 IEEE/CVF Conference on Computer Vision and Pattern Recognition Workshops (CVPRW)}} (\bibinfo{year}{2018}), \bibinfo{pages}{398--407}.
\newblock
\urldef\tempurl%
\url{https://api.semanticscholar.org/CorpusID:4797043}
\showURL{%
\tempurl}


\bibitem[Zhao and Wu(2019)]%
        {SaliencyModern}
\bibfield{author}{\bibinfo{person}{Ting Zhao} {and} \bibinfo{person}{Xiangqian Wu}.} \bibinfo{year}{2019}\natexlab{}.
\newblock \showarticletitle{Pyramid Feature Attention Network for Saliency Detection}. In \bibinfo{booktitle}{\emph{2019 IEEE/CVF Conference on Computer Vision and Pattern Recognition (CVPR)}}. \bibinfo{pages}{3080--3089}.
\newblock
\urldef\tempurl%
\url{https://doi.org/10.1109/CVPR.2019.00320}
\showDOI{\tempurl}


\bibitem[Zhou et~al\mbox{.}(2016)]%
        {cam}
\bibfield{author}{\bibinfo{person}{Bolei Zhou}, \bibinfo{person}{Aditya Khosla}, \bibinfo{person}{Agata Lapedriza}, \bibinfo{person}{Aude Oliva}, {and} \bibinfo{person}{Antonio Torralba}.} \bibinfo{year}{2016}\natexlab{}.
\newblock \showarticletitle{Learning Deep Features for Discriminative Localization}. In \bibinfo{booktitle}{\emph{Proceedings of the IEEE Conference on Computer Vision and Pattern Recognition (CVPR)}}.
\newblock


\bibitem[Zou et~al\mbox{.}(2023)]%
        {SegmentationSEEM}
\bibfield{author}{\bibinfo{person}{Xueyan Zou}, \bibinfo{person}{Jianwei Yang}, \bibinfo{person}{Hao Zhang}, \bibinfo{person}{Feng Li}, \bibinfo{person}{Linjie Li}, \bibinfo{person}{Jianfeng Wang}, \bibinfo{person}{Lijuan Wang}, \bibinfo{person}{Jianfeng Gao}, {and} \bibinfo{person}{Yong~Jae Lee}.} \bibinfo{year}{2023}\natexlab{}.
\newblock \bibinfo{title}{Segment Everything Everywhere All at Once}.
\newblock
\newblock
\showeprint[arxiv]{2304.06718}~[cs.CV]


\end{thebibliography}

\appendix

\end{document}